\documentclass[acmlarge, nonacm]{acmart}

\AtBeginDocument{%
  \providecommand\BibTeX{{%
    \normalfont B\kern-0.5em{\scshape i\kern-0.25em b}\kern-0.8em\TeX}}}

\usepackage{amsmath}
 
\usepackage{balance}       
\usepackage{graphics}      
\usepackage[T1]{fontenc}   
\usepackage{lipsum}
\usepackage{color}
\usepackage{booktabs}
\usepackage{textcomp}
\usepackage{microtype}        
\usepackage{ccicons}          
\usepackage{multirow}  
\usepackage{color,soul}  
\usepackage{subcaption}  
\usepackage{listings}  
\usepackage{todonotes}

\definecolor{ao(english)}{rgb}{0.0, 0.5, 0.0}
\newcommand{\harish}[1]{\textbf{\sffamily{\textcolor{ao(english)}{[#1 -- Harish]}}}}

\def\etal{\emph{et al}.}

\usepackage{fancyhdr}   
\fancypagestyle{firststyle}{%
  \fancyhf{}
  \fancyfoot[C]{\textcolor{red}{This manuscript is under review. Please write to harishkashyap@gatech.edu for up-to-date information.}}
}
\fancypagestyle{allstyle}{%
   \fancyfoot{}
   \fancyfoot[C]{\textcolor{red}{This manuscript is under review. Please write to harishkashyap@gatech.edu for up-to-date information.}}
}

\setcopyright{acmcopyright}
\acmJournal{IMWUT}
\acmYear{2020} 
\acmVolume{4} 
\acmNumber{42} 
\acmArticle{0815} 
\acmMonth{11}
\acmPrice{15.00}
\acmDOI{XXXXXXXXXXXX}

\begin{document}

\title{Contrastive Predictive Coding for Human Activity Recognition}
\lfoot{\vfill \textcolor{red}{This manuscript is under review. Please contact harishkashyap@gatech.edu for up-to-date information.}}

\author{Harish Haresamudram}
\email{hharesamudram3@gatech.edu}
\affiliation{%
	\institution{School of Electrical and Computer Engineering, Georgia Institute of Technology}
	\city{Atlanta, GA}
	\country{USA}
}

\author{Irfan Essa}
\email{irfan@gatech.edu}
\affiliation{
	\institution{School of Interactive Computing, Georgia Institute of Technology}
	\city{Atlanta, GA}
	\country{USA}
}

\author{Thomas Pl\"{o}tz}
\email{thomas.ploetz@gatech.edu}
\affiliation{
	\institution{School of Interactive Computing, Georgia Institute of Technology}
	\city{Atlanta, GA}
	\country{USA}
}

\renewcommand{\shortauthors}{Haresamudram, et al.}

\begin{abstract}
Feature extraction is crucial for human activity recognition (HAR) using body-worn movement sensors.
Recently, learned representations have been used successfully, offering promising alternatives to manually engineered features.
Our work focuses on effective use of small amounts of labeled data and the opportunistic exploitation of unlabeled data that are straightforward to collect in mobile and ubiquitous computing scenarios.
We hypothesize and demonstrate that explicitly considering the temporality of sensor data at representation level plays an important role for effective HAR in challenging scenarios.
We introduce the Contrastive Predictive Coding (CPC) framework to human activity recognition, which captures the long-term temporal structure of sensor data streams. 
Through a range of experimental evaluations on real-life recognition tasks, we demonstrate its effectiveness for improved HAR. 
CPC-based pre-training is self-supervised, and the resulting learned representations can be integrated into standard activity chains.
It leads to significantly improved recognition performance when only small amounts of labeled training data are available, thereby demonstrating the practical value of our approach.
\end{abstract}




\maketitle
\thispagestyle{firststyle} 
\pagestyle{allstyle}

\section{Introduction}
Body-worn movement sensors, such as accelerometers or full-fledged inertial measurement units (IMU), have been extensively utilized for a wide range of applications in mobile and ubiquitous computing, including but not limited to novel interaction paradigms \cite{zhang2016tapskin, reyes2016whoosh, zhang2017soundtrak}, gesture recognition \cite{zhang2017fingersound}, eating detection~\cite{edison_eating, bedri2017earbit, amft2005detection, zhang2020necksense}, and health and well-being assessments in general \cite{morshed2019prediction, wang2014studentlife, fisher2016unsupervised}. 
They are widely utilized on commodity smartphones, and smartwatches such as Fitbit and the Apple Watch.
The ubiquitous nature of these devices makes them highly suitable for real-time capturing and analysis of activities as they are being performed. 

The workflow for human activity recognition (HAR), i.e., the all encompassing paradigm for aforementioned applications, essentially involves the recording of movement data after which signal processing and machine learning techniques are applied to automatically recognize the activities. 
This type of workflow is typically supervised in nature, i.e., it requires the labeling of what activities have been performed and when after the data collection is complete \cite{bulling2014tutorial}. 
Streams of sensor data are segmented into individual analysis frames using a sliding window approach, and forwarded as input into feature extractors. 
The resulting representations are then categorized by a machine learning based classification backend into the activities under study (or the NULL class).

The availability of large-scale annotated datasets has resulted in astonishing improvements in performance due to the application of deep learning to computer vision \cite{krizhevsky2017imagenet, he2016deep}, speech recognition \cite{graves2013speech, amodei2016deep} and natural language tasks \cite{mikolov2013distributed, devlin2018bert}. 
While end-to-end training has also been applied to activity recognition from wearable sensors~\cite{guan2017ensembles, hammerla2016deep, ordonez2016deep}, the depth and complexity is limited by a lack of such large-scale, diverse labeled data. 
However, due to the ubiquity of sensors (e.g., in phones and commercially available wearables such as watches etc.) the data recording itself is typically straightforward, which is in contrast to obtaining their annotations, thereby resulting in potentially large quantities of unlabeled data.
Thus, in our work we look for approaches that can make economic use of the limited labeled data and exploit unlabeled data as effectively as possible. 

Previous works such as \cite{haresamudram2019role, saeed2019multi, plotz2011feature} have demonstrated how unlabeled data can be utilized to learn useful representations for wide ranging tasks, including identifying kitchen activities \cite{chavarriaga2013opportunity}, activity tracking in car manufacturing \cite{stiefmeier2008wearable}, classifying every day activities such as walking or running \cite{reiss2012introducing, anguita2013public, chatzaki2016human, malekzadeh2018protecting}, and medical scenarios involving identifying freeze of gait in patients suffering from Parkinson's disease \cite{nutt2011freezing}. 
In many such applications, the presence of complex and often sparsely occurring movement patterns coupled with limited annotation makes it especially hard for deriving effective recognition systems. 
The promising results delivered in these works without the use of labels have resulted in a general direction of integrating unsupervised learning as-is into conventional activity recognition chains (ARC) \cite{bulling2014tutorial} in the feature extraction step. 
In this work, we follow this general direction of utilizing (potentially large amounts of) unlabeled data for effective representation learning and subsequently construct activity recognizers from the representations learned. 

Recent work towards such unsupervised pre-training has gone beyond the early introduction using Restricted Boltzmann Machines (RBMs) \cite{plotz2011feature}, involving (variants of) autoencoders \cite{haresamudram2019role, varamin2018deep}, and self-supervision \cite{saeed2019multi, haresamudram2020masked}.
While they result in effective representations, most of these approaches do not specifically target a characteristic inherent to body-worn sensor data -- temporality.
Wearable sensor data resemble time-series and we hypothesize that incorporating temporal characteristics directly at the representation learning level results in more discriminative features and more effective modeling, thereby leading to better recognition accuracy for HAR scenarios with limited availability of labeled training data -- as they are typical for mobile and ubiquitous computing scenarios.

Previous work on masked reconstruction \cite{haresamudram2020masked} has attempted to address temporality at feature level in a self-supervised learning scenario by regressing to the zeroed sensor data at randomly chosen timesteps. 
This incorporates local temporal characteristics into a pretext task that forces the recognition network to predict missing values based on immediate past and future data. 
It was shown that the resulting sensor data representations are beneficial for modeling activities, which provides evidence for our aforementioned hypothesis of temporality at feature level playing a key role for effective modeling in HAR under challenging constraints.

In this paper we present a framework that rigorously follows the paradigm of modeling temporality at representation level. 
We propose to utilize Contrastive Predictive Coding (CPC) \cite{oord2018representation} for unsupervised representation learning of windowed body-worn sensor data. 
The key insight behind this approach is that predicting just the next future timestep involves exploiting the local smoothness of the signal, whereas predicting multiple subsequent timesteps requires inferring the global structure of time-series data.
This results in representations that encode the high-level information between temporally separated parts of the time-series signal \cite{oord2018representation}, thereby leading to improved downstream recognition performance. 
Sliding windows of movement data are passed through a non-linear encoder to map the data into a latent space, where an autoregressive network such as gated recurrent units (GRUs) is subsequently utilized to provide the context representation. 
Using this context, multiple subsequent future timesteps from the encoder are predicted and noise contrastive estimation (NCE) \cite{gutmann2010noise} is utilized to train the model.
Pre-trained weights are subsequently used in the Activity Recognition Chain at the feature extraction step.
We apply our framework for learned representations of activities on four benchmark datasets and demonstrate improved recognition performance.
In summary, our contributions are as follows:
\begin{itemize}
	\item We introduce Contrastive Predictive Coding as an effective self-supervised pre-training scheme for HAR.
	\item Through a series of experiments, we analyze the network architecture choices and optimal training settings.
	\item We demonstrate economic use of available annotations by fine-tuning pre-trained models with limited labeled data, resulting in significant improvements over end-to-end training on the available annotations.
\end{itemize}

\section{Related Work on Representations of Sensor Data in Human Activity Recognition}
In our work, we focus on learning effective representations for movement data recorded from body-worn sensing platforms through unsupervised pre-training. 
This follows the general direction of reducing reliance on annotated data, by utilizing approaches that perform learning on unlabeled data. 
As such, relevant prior works involve the following: 
\emph{(i)} feature learning for activity recognition in general;
\emph{(ii)} self-supervised learning in general, i.e., not limited to the mobile and ubiquitous computing domain; and 
\emph{(iii)} self-supervised learning for HAR.

The activity recognition chain (ARC) \cite{bulling2014tutorial} details a five step pipeline for human activity recognition.
We are interested in the fourth step in the process -- feature extraction -- which focuses on the computation of representative features.
Earlier works utilized hand crafted features for activity recognition \cite{figo2010preprocessing}.
Recently, however, end-to-end training has been increasingly adopted for HAR since it offers integrated representation learning capabilities. 
There is not yet a consensus on the gold standard feature representation for HAR using wearables \cite{haresamudram2019role}.
However, they can broadly be broken down into three categories: 
\emph{(i)} Statistical features, e.g., mean and standard deviation of raw data \cite{huynh2005analyzing}; 
\emph{(ii)} Distribution-based representations, including those based on empirical cumulative distribution functions of the raw data \cite{hammerla2013preserving}; and
\emph{(iii)} Learned features, that directly utilize the data itself to derive representations, involving supervised or unsupervised learning and dimensionality reduction techniques~\cite{plotz2011feature, ordonez2016deep, haresamudram2019role}. 

\subsection{Feature Engineering}
More traditional approaches aim at finding compact representations for the sensor data. 
A wealth of heuristics, both in the time and frequency domain, have been developed towards extracting such representations~\cite{figo2010preprocessing}.
The distribution-based representations, involving the quantiles of the (inverted) empirical cumulative density function of sensor windows present the state-of-the-art for conventional feature extraction \cite{hammerla2013preserving}.
Improvements to this representation technique include adding structure \cite{kwon2018adding} and specializing window lengths for human activity recognition \cite{li2018specialized}. 
One type of learned features includes dimensionality reduction techniques such as the principal component analysis, which has been utilized as a feature for activity recognition \cite{plotz2011feature}. 

\subsection{Feature Learning}
Methods for feature learning aim at optimizing dedicated objective functions in order to learn useful representations from raw sensor data.
Supervised learning requires annotated datasets and does not explicitly differentiate between the representation learning and classification steps of the ARC. 
As such, feature extraction (learning) is performed implicitly by the particular model that is being trained.
Convolutional networks have been utilized to classify multi-variate sensor data~\cite{yang2015deep, zeng2014convolutional, hammerla2016deep}. 
Leveraging the temporal correlations in wearable sensor data at modeling level, (variants of) recurrent neural networks (RNNs) have been applied previously \cite{guan2017ensembles, hammerla2016deep, zeng2018understanding}.
Guan \etal~\cite{guan2017ensembles} setup ensemble training with recurrent networks in order to improve classification performance. 
Continuous attention mechanisms over both the sensory and time channels have been applied in \cite{zeng2018understanding} in order to learn both the `important' timesteps as well as channels in windows of data. 
A combination of both convolutional and recurrent networks is utilized by Ordonez \etal \ \cite{ordonez2016deep} where 2D convolutional features are fed to a long short-term memory \cite{hochreiter1997long} network to perform activity recognition. 
Subsequent work on this architecture to improve performance includes temporal attention \cite{murahari2018attention}. 
More recently, Transformer \cite{vaswani2017attention} encoder networks have been utilized for activity recognition \cite{mahmud2020human, haresamudram2020masked}.
They model sequential information via the use of self-attention mechanisms and utilize only dense layers.
As they dispense of both convolutions and recurrence, they are more parallelizable and require less time to train \cite{vaswani2017attention}.
Overall, supervised learning remains the de facto approach towards activity recognition in recent years and also comprises the state-of-the-art. 

Prior work into unsupervised learning has involved RBMs \cite{plotz2011feature} and variants of autoencoder models \cite{haresamudram2019role, varamin2018deep}.
In \cite{haresamudram2019role} the authors investigate the role of representations in activity recognition, by evaluating different features on a common classification backend. 
Varamin \etal~\cite{varamin2018deep} define HAR as a set prediction problems within the autoencoder setup. 
A more recent development includes the use of self-supervised learning for unsupervised pre-training. 
Along with Haresamudram \etal`s convolutional autoencoder (CAE) \cite{haresamudram2019role}, these self-supervised learning approaches form our unsupervised baselines.

\subsection{Self-Supervised Learning}
Self-supervised learning involves defining a `pretext' task from the data itself, that provides supervisory signals beneficial for downstream tasks. 
Numerous such tasks have been designed for both computer vision and natural language processing problems.
For example, colorization \cite{zhang2016colorful} involves a model that is trained to colorize grayscale images. 
Predicting image rotations \cite{gidaris2018unsupervised} allows the model to learn concepts of the objects in the images, such as their location, pose and type. 
This approach is shown to be beneficial towards a series of downstream  tasks including classification and object recognition. 
Other approaches such as in-painting \cite{pathak2016context} mask portions of the image, and predicting relative positions of patches \cite{doersch2015unsupervised} have also been shown to perform well for representation learning.
Self-supervision involving spatio-temporality has been effectively applied for learning representations for videos. 
Identifying the odd or unrelated video sub-sequence \cite{fernando2017self}, sequence verification of temporal order  \cite{misra2016shuffle}, and learning the arrow of time \cite{wei2018learning} have been used as pretext tasks. 

In the case of natural language processing, context-based self-supervised learning has been applied to obtain word and sentence level embeddings for approaches such as Word2Vec \cite{mikolov2013distributed}, GloVe \cite{pennington2014glove} and universal sentence encoder \cite{cer2018universal}.
Most recently, transformer-based \cite{vaswani2017attention} approaches such as GPT \cite{radford2018improving} and BERT \cite{devlin2018bert} have been developed.
BERT involves predicting masked tokens from surrounding context followed by predicting the next sentence. 
GPT similarly learns useful representations by pre-training a language model and then discriminatively fine-tuning on each specific task. 

A more related field to activity recognition from wearable sensing involves speech and audio processing.
In \cite{wang2020unsupervised}, Wang \etal\ perform self-supervision by reconstructing randomly masked out groups of timesteps as well as frequencies on log mel filterbank energies by using bidirectional recurrent encoders.
In a similar vein, Liu \etal \cite{liu2020tera} utilize reconstruction of time, channel and magnitude altered inputs for self-supervision (see also 
\cite{liu2020mockingjay, zhao2020musicoder}.) 
Autoregressive predictive coding has also been explored for speech self-supervision in \cite{chung2019unsupervised, chung2020generative} and \cite{chung2020improved}.
Another major family of self-supervised learning approaches involves contrastive learning. 
The intuition behind contrastive learning is to compare semantically similar (positive) and dissimilar (negative) pairs of data points, and to encourage the distance between similar pairs to be close while the distance between dissimilar pairs is encouraged to be more orthogonal \cite{chuang2020debiased}. 
This framework has resulted in excellent representation learning performance across domains such as computer vision \cite{chen2020simple, he2020momentum, henaff2019data}, natural language \cite{logeswaran2018efficient}, speech recognition \cite{baevski2020wav2vec, oord2018representation} and time-series data \cite{hyvarinen2016unsupervised}.
For example, Schneider \etal~\cite{baevski2020wav2vec} utilize a contrastive loss on raw speech signals from a large unlabeled corpus to improve speech recognition performance on smaller labeled datasets. 
Contrastive Predictive Coding (CPC) \cite{oord2018representation}, which is the focus of this work, performs contrastive learning over multiple future timestep predictions. 

\subsection{Self-Supervised Learning for HAR}
\label{sec:rw_self_sup_HAR}
Self-supervision for activity recognition from wearables has been explored from a multi-task learning context. 
In \cite{saeed2019multi}, eight data transformations were applied separately to the input data.
These transformations were applied randomly, i.e. with $50\%$ probability, and the task consisted of identifying if each transformation was applied or not in a multi-task setting. 
The architecture involved a transformation prediction network (TPN) consisting of 1D convolutional layers common to all tasks along with task specific fully connected layers. 
The learned TPN weights were used for representation learning, transfer learning and semi-supervised learning with limited labeled data.
Multi-task self-supervised learning \cite{saeed2019multi} utilizes accelerometer signals while our work is agnostic to sensor modalities.
Therefore, it is applicable to a wider range of scenarios and environments. 

A more recent work involves the use of Transformer \cite{vaswani2017attention} encoders to reconstruct randomly masked timesteps in windows of sensor data \cite{haresamudram2020masked}.
Here, sensor data at $10\%$ of randomly chosen timesteps is masked, or set to zero.
The transformer encoder layers are subsequently trained to reconstruct only the missing data and learned weights are used for activity recognition, transfer learning and fine-tuning with limited labeled data.
While \cite{haresamudram2020masked} utilizes both accelerometer and gyroscope data, it delivers mixed results relative to DeepConvLSTM \cite{ordonez2016deep} when fine-tuning the pre-trained weights using limited labeled data. 
In comparison, our work results in consistent improvement in performance on all benchmark datasets over DeepConvLSTM and thereby reduces reliance on the quantity of labeled data required for training systems. 

\section{Self-supervised pre-training with Contrastive Predictive Coding}
\begin{figure*}[t]
	\centering
	\includegraphics[width=0.95\textwidth]{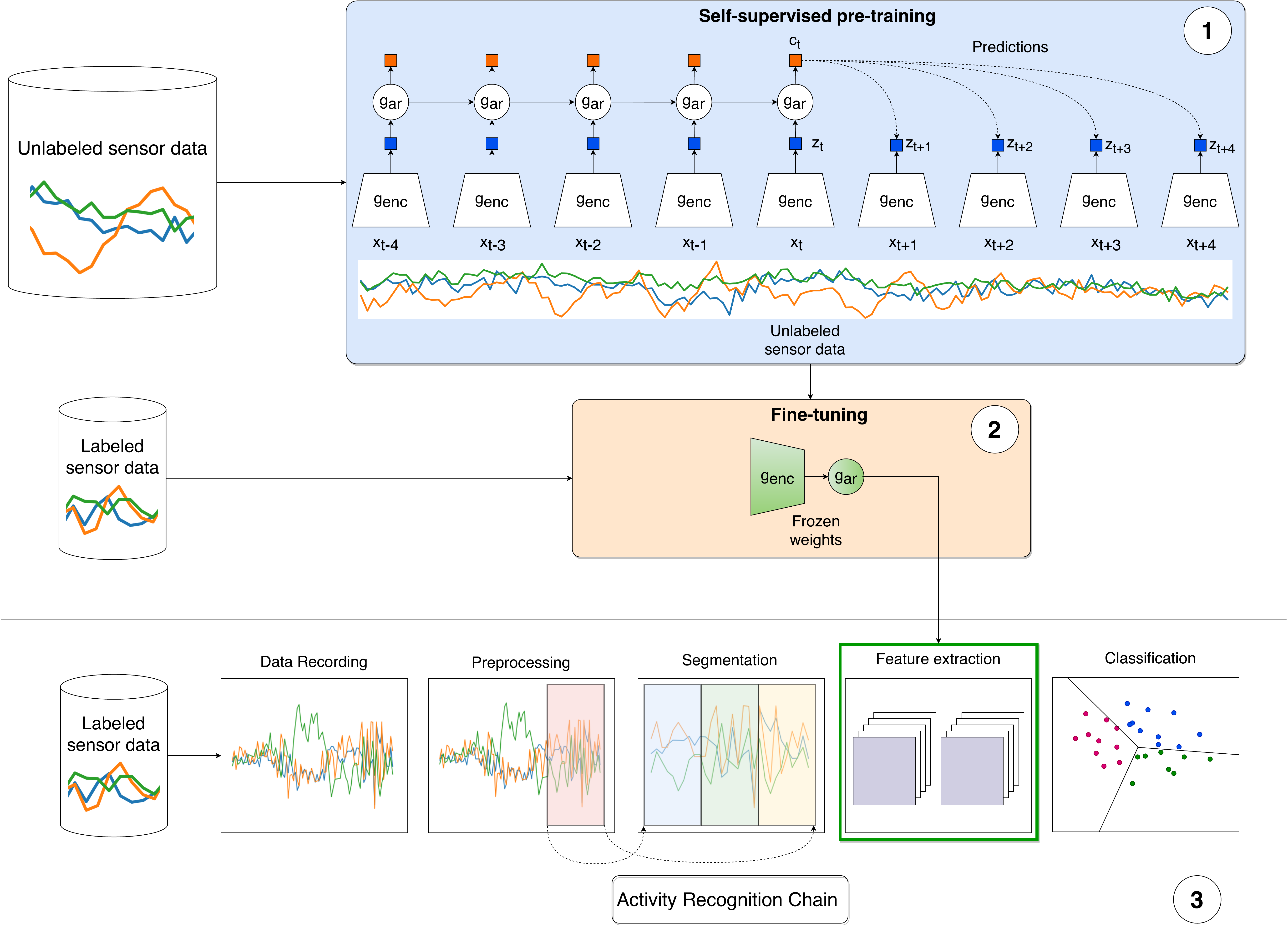}
	\caption{Overview of proposed Human Activity Recognition framework based on Contrastive Predictive Coding (CPC) -- details explained in main text.}
	\label{fig:cpc}
\end{figure*}

\noindent
In this paper, we introduce the Contrastive Predictive Coding (CPC) framework to human activity recognition from wearables.
Fig.\ \ref{fig:cpc} outlines the overall workflow, which includes:
\emph{(i) pre-training} (part 1 in Fig.\ \ref{fig:cpc}), where unlabeled data are utilized to obtain useful representations (i.e., learn encoder weights) via the pretext task; and, 
\emph{(ii) fine-tuning}, which involves performing activity recognition on the learned representations using a classifier (part 2 in Fig.\ \ref{fig:cpc}). 
During pre-training, the sliding window approach is applied to large quantities of unlabeled data to segment it into overlapping windows.
They are utilized as input for self-supervised pre-training, which learns useful unsupervised representations. 
Once the pre-training is complete, weights from both $g_{enc}$ and $g_{ar}$ are frozen and used for feature extraction (part 2 in Fig.\ \ref{fig:cpc}).
This corresponds to the feature extraction step in the ARC (part 3 in Fig.\ \ref{fig:cpc}).

The frozen learned weights are utilized with the backend classifier network (see Sec.\ \ref{sec:backend_classifier}), a three-layer multi-layer perceptron (MLP), in order to classify windows of labeled data into activities.
This corresponds to the classification step in the ARC.
The learned weights from CPC are frozen and only the classifier is optimized on (potentially smaller amounts of) labeled datasets.
The resulting performance directly indicates the quality of the learned representations.

In what follows, we first detail our Contrastive Predictive Coding framework as it is applied to HAR, and then describe the backend classifier network used to evaluate the unsupervised representations.

\subsection{Contrastive Predictive Coding}
\label{sec:cpc}
Contrastive predictive coding (CPC) involves predicting future timesteps in the latent space using autoregressive modeling. 
The intuition behind such a task is to extract high-level information from the signal while discarding the more local noises \cite{oord2018representation}.
Predicting one future timestep is an established approach in signal processing \cite{elias1955predictive, atal1970adaptive} and exploits the local smoothness of signals. 
However, as the model is forced to predict farther into the future, it needs to infer more global structure \cite{oord2018representation}. 
This results in the incorporation of long-term temporal characteristics into the representation learning itself.
We hypothesize that this process is beneficial towards learning effective representations.

The architecture for CPC is detailed in part 1 of Fig.\ \ref{fig:cpc}. 
As detailed in \cite{oord2018representation}, a non-linear encoder $g_{enc}$ is utilized to map a window of sensor data $x_t$ to a sequence of latent representations given by $z_t=g_{enc}(x_t)$. 
Subsequently, an autoregressive model $g_{ar}$ is used to summarize all $z_{\leq t}$ into a latent space, providing a context latent representation $c_t = g_{ar}(z_{\leq t})$. 
%
Instead of predicting the future samples $x_{t+k}$ directly using a generative model, a density ratio which preserves the mutual information between $x_{t+k}$ and $c_t$ is modeled as follows:
\begin{equation}
	f_k(x_{t+k}, c_t) \propto \frac{p(x_{t+k}|c_t)}{p(x_{t+k})}.
\end{equation}
The density ratio $f$ is unnormalized and a simple log-bilinear model is used for scoring:
\begin{equation}
	f_k(x_{t+k}, c_t) = exp(z^{T}_{t+k}W_{k}c_{t})
\end{equation}

A linear transformation $W_{k}c_{t}$ is used to predict the $k$ timesteps, with a separate $W_{k}$ being used for each step. 
Using the density ratio $f_k(x_{t+k}, c_t)$ and inferring $z_{t+k}$ using an encoder relieves the network from modeling the high dimensional distribution $x_{t_k}$ and instead allows for sampling $p(x)$ or $p(x|c)$ directly using Noise Contrastive Estimation (NCE). 
We utilize the InfoNCE loss detailed in \cite{oord2018representation} to update the network parameters:
\begin{equation}
	L_N = -\mathbb{E}_X \left[ log \frac{f_k(x_{t+k}, c_t)} {\sum_{x_j \in X}^{} f_k(x_j, c_t)}   \right]
	\label{eqn:info_nce}
\end{equation}

Here, one positive sample from $p(x_{t+k}|c_t)$ and $N-1$ negative samples from the `proposal' distribution $p(x_{t+k})$ are utilized for optimizing the loss. 
The pre-training using Eq.\ \ref{eqn:info_nce} involves classifying the positive sample correctly from a combined set of one positive and $N-1$ negative samples. 
After the pre-training is complete, we utilize $g_{enc}$ and $g_{ar}$ weights to extract the representations.

From an implementation standpoint, given a window of $T$ timesteps and $k$ future step predictions, we pass the entire window through the encoder.
We choose a random timestep $t \in [0, T-k]$ and predict $k$ subsequent timesteps. 
Sensor data present between $0$ and $t$ are utilized as input to the autoregressive network $g_{ar}$, which results in the context representation $c_t$.
For each future timestep prediction, the negatives comprise of same timestep from all other windows in the batch.

\subsection{Backend Classifier Network}
\label{sec:backend_classifier}
In Sec.\ \ref{sec:cpc}, we detailed the contrastive predictive coding framework, which constitutes the Feature Extraction step in the ARC.
The subsequent step in the chain -- Classification -- involves the evaluation of the representations learned via self-supervision.
The classification backend consists of a $3$ layer feedforward network with the first two layers comprising of $256$ and $128$ units respectively.
The last layer constitutes the softmax layer, the size of which equals the number of classes in the dataset.
Between the layers, we apply batch normalization \cite{ioffe2015batch}, the ReLU activation function \cite{nair2010rectified} and dropout \cite{srivastava2014dropout} with p=$0.2$. 
This network is identical to the classifier used in Haresamudram \etal \cite{haresamudram2020masked}, thereby making the results obtained directly comparable. 


\section{Human Activity Recognition based on Contrastive Predictive Coding}
In the previous section we have introduced our representation learning framework for movement data based on contrastive predictive coding.
This pre-training step is integrated into an overarching human activity recognition framework, that is based on the standard Activity Recognition Chain (ARC) \cite{bulling2014tutorial}.
Addressing our general goal of deriving effective HAR systems from limited amounts of annotated training data, as it is a regular challenge in mobile and ubiquitous computing settings, we conducted extensive experimental evaluations to explore the overall effectiveness of our proposed representation learning approach.

In what follows we provide a detailed explanation of our experimental evaluation, which includes descriptions of: 
\textit{i)} Application scenarios that our work focuses on;
\textit{ii)} Implementation details;
\textit{iii)} Evaluation metrics used for quantitative evaluation; and
\textit{iv)} Overall experimental procedure.
Results of our experiments and discussion thereof are presented in Sec.\ \ref{sec:results}.

\subsection{Application Scenarios}
\label{sec:datasets}
In our work, we focus on studying self-supervised pre-training primarily for locomotion style activities such as walking, running, and sitting, as these broadly represent a significant portion of activities performed daily and are of significant interest for the mobile and ubiquitous computing community \cite{Abowd:2012vt}.
The effectiveness of our CPC-based pre-training is evaluated on four representative benchmark datasets, which mainly contain data recorded from smartphones (see Tab.\ \ref{tab:datasets} for an overview). 
In particular, recording movement data from smartphones has the advantage of being unobtrusive, inexpensive, and available to large portions of the population.
This also allows for easier in-the-wild data collection. 
The chosen datasets cover diverse participants, environments of study, and data collection protocols for body-worn sensors including accelerometers and gyroscopes and as such are representative for the targeted application scenarios.
They were also studied in previous works on self-supervised learning in human activity recognition \cite{haresamudram2020masked, saeed2019multi}. 
Additionally, we include the USC-HAD dataset into our explorations, as an example for an activity recognition scenario using body-worn sensing platforms that are not smartphones.

Unless specified differently, dataset splits for training, validation and testing are performed based on participant IDs.
$20\%$ of the participants are randomly chosen to form the test dataset. 
Of the remaining data, $20\%$ are randomly chosen once again to form the validation set, while the rest are utilized for training. 
Detailed descriptions of the datasets are given below.

\begin{table}[t]
	\centering
	\caption{Overview of the datasets used in the evaluation. 
	}
	\begin{tabular}{c|c|c|c|c}
		\hline
		Dataset & \begin{tabular}[c]{@{}c@{}}\# of\\  users\end{tabular} & \begin{tabular}[c]{@{}c@{}}\# Activity \\ classes\end{tabular} & \begin{tabular}[c]{@{}c@{}}Recording \\ Device\end{tabular} & Activities \\ \hline
		Mobiact & 61 & 11 & \begin{tabular}[c]{@{}c@{}}Samsung \\ Galaxy S3\end{tabular} & \begin{tabular}[c]{@{}c@{}}Sitting, walking, jogging, jumping, \\ stairs up, stairs down, stand to sit, sitting \\ on a chair,  sit to stand, car step-in, and car step-out\end{tabular} \\ \hline
		Motionsense & 24 & 6 & iPhone 6S & \begin{tabular}[c]{@{}c@{}}Walking, jogging, going up and down \\ the stairs, sitting and standing\end{tabular} \\ \hline
		UCI-HAR & 30 & 6 & \begin{tabular}[c]{@{}c@{}}Samsung \\ Galaxy S2\end{tabular} & \begin{tabular}[c]{@{}c@{}}Standing, sitting, walking, lying down, \\ walking downstairs and upstairs\end{tabular} \\ \hline
		USC-HAD & 14 & 12 & \begin{tabular}[c]{@{}c@{}}MotionNode \\ platform\end{tabular} & \begin{tabular}[c]{@{}c@{}}Walking - forward, left, right, upstairs, and downstairs, \\ running forward, jumping, sitting, standing, sleeping, \\ and riding the elevator up and down\end{tabular} \\ \hline
	\end{tabular}
	\label{tab:datasets}
\end{table}

\subsubsection{Mobiact}
Movement data from inertial measurement units were collected using a Samsung Galaxy S3 smartphone placed freely in a trouser pocket.
The dataset covers eleven activities of daily living and four types of falls \cite{chatzaki2016human}. 
We used v$2$ of the dataset and subset data that cover only daily living activities, resulting in a total of $61$ participants.
The activities include: sitting, walking, jogging, jumping, stairs up, stairs down, stand to sit, sitting on a chair, sit to stand, car step-in, and car step-out.

\subsubsection{Motionsense}
The Motionsense dataset \cite{malekzadeh2018protecting} consists of $24$ participants of different ages, gender, weight and height. 
The dataset was collected to propose a representation learning model that offers flexible and negotiable privacy-preserving sensor data transmission. 
The data were collected using an iPhone $6$s and comprises accelerometer, gyroscope and attitude information.
The activities covered include walking, jogging, going up and down the stairs, sitting and standing. 

\subsubsection{UCI-HAR}
The UCI-HAR dataset \cite{anguita2013public} contains data collected from $30$ participants using a waist-mounted Samsung Galaxy S$2$ smartphone.
The subjects performed six activities including standing, sitting, walking, lying down, walking downstairs and upstairs (the transition classes are not included). 
We use raw data from both the accelerometer and gyroscope in our study (also called the HAPT dataset \footnote{http://archive.ics.uci.edu/ml/datasets/Smartphone-Based+Recognition+of+Human+Activities+and+Postural+Transitions}). 

\subsubsection{USC-HAD}
The USC-HAD dataset \cite{zhang2012usc} was collected on the MotionNode sensing platform and consists of data from $14$ subjects. 
Twelve activities were recorded, including walking--forward, left, right, upstairs, and downstairs--, running forward, jumping, sitting, standing, sleeping, and riding the elevator up and down.
Following the protocol from Haresamudram \etal\  \cite{haresamudram2019role}, participants $1-10$ form the training set, while participants $11$ and $12$ form the validation set, and participants $13$ and $14$ comprise the test set.

\subsection{Implementation Details}
We implemented our framework using PyTorch \cite{paszke2019pytorch}.
Source code will be shared when the paper is published.
In what follows we provide details of parameter choices for the overall processing framework, which shall allow the reader to replicate our experiments.

\subsubsection{Data Preparation}
We use raw accelerometer and gyroscope data from the benchmark datasets detailed in Sec.\ \ref{sec:datasets}.
As deep networks can effectively learn abstract representations from raw data itself, we perform no further filtering / denoising on the datasets \cite{lecun2015deep}. 
The datasets have different sampling rates and we downsample them to $30$Hz in order to maintain uniformity.
Further, we also normalize the individual channels of the train dataset split to zero mean and unit variance.
These means and variances are subsequently applied to the validation and test splits. 
Similar to prior works, sliding window based segmentation is applied to obtain windows of $1$s with $50\%$ overlap between subsequent windows \cite{guan2017ensembles, hammerla2016deep, haresamudram2020masked}.

\subsubsection{CPC Pre-Training}
The pre-training is performed for $150$ epochs with the learning rate being tuned over $\{1e-3, 5e-4\}$ and $k \in \{2, 4, 8, 12, 16\}$. 
The network weights are optimized using the Adam \cite{kingma2014adam} optimizer.
For the $g_{enc}$ network, we primarily evaluate a 1D Convolutional Encoder, containing three blocks with 1D convolutional layers of 32, 64 and 128 channels respectively with a kernel size of $3$. 
Each block consists of a 1D convolutional layer with reflect padding, followed by the ReLU activation function and dropout with p=$0.2$. 
For the autoregressive network $g_{ar}$, we utilize a two-layer gated recurrent units (GRU) network \cite{chung2014empirical} containing 256 units and dropout with p=$0.2$.
The prediction networks, $W_k$, are linear layers with 128 units.  

\subsubsection{Activity Recognition}
The classification backend is trained with labeled data for $150$ epochs using cross entropy loss.
Learning rate is tuned over $\{5e-4, 1e-4\}$ and is decayed by a factor of $0.8$ every $25$ epochs.
The network parameters are updated using the Adam \cite{kingma2014adam} optimizer. 

\subsection{Performance Metric}
The test set mean F1-score is utilized as the primary metric to evaluate performance. 
The datasets used in this study show substantial class imbalance and thus experiments require evaluation metrics that are less affected negatively by such biased class distributions \cite{powers2020evaluation}.
The mean F1-score is given by:

\begin{equation}
	F_m = \frac{2}{|c|}\sum_{c}^{} \frac{prec_{c} \times recall_{c}}{prec_{c} + recall_{c}}
\end{equation}
where $|c|$ corresponds to the number of classes while $prec_c$ and $recall_c$ are the precision and recall for each class. 
\subsection{Experimental Procedure}
As the main focus of this work is on deriving effective HAR representations without relying on labels, we study the self-supervised pre-training from two perspectives:
\begin{enumerate}
\item First, we compute the activity recognition performance, which describes the quality of the representations learned during pre-training. 
It indicates the raw ability of the representations to be discriminative towards the activities under study.
\item Then, we extend this evaluation to scenarios with limited availability of annotated data and compare performance of the learned weights to end-to-end training.
This considers practical situations where very limited labeling is possible from users after the deployment of recognition systems.
\end{enumerate}
Put together, these evaluations provide a rounded understanding of the effectiveness of the self-supervised pre-training and its performance for representation learning.

\section{Results and Discussion}
\label{sec:results}
\subsection{Activity Recognition} 
\label{sec:rep_learning}

\begin{table}[t]
	\centering
	\caption{
		We compare the representation learning performance of the proposed approach against both supervised learning and to state-of-the-art unsupervised learning baselines. 
		Performance for the approaches with $\dagger$ have been taken from \cite{haresamudram2020masked}.
		CPC (end-to-end) refers to the setup wherein the same network architecture as CPC (1D Conv Encoder) is utilized.
		However, all weights of the network are initialized randomly and trained end-to-end from scratch using labeled data.
		Results marked in \textbf{\textcolor{green}{green}} correspond to the best performing models including both supervised and unsupervised learning approaches.
		For easy comparison, results in \textbf{bold} include the best performing unsupervised learning techniques.
	}
	\begin{tabular}{c|c|c|c|c|c}
		\hline
		Approach & Method type & Mobiact & Motionsense & UCI-HAR & USC-HAD \\ \hline
		DeepConvLSTM$^{\dagger}$ \cite{ordonez2016deep}& Supervised & 82.40 & 85.15 & \textbf{\textcolor{green}{82.83}} & 44.83 \\ 
		CPC (end-to-end, 1D Conv Encoder) & Supervised & \textbf{\textcolor{green}{83.68}} & 86.66 & 79.79 & 49.09 \\ \hline \hline
		Multi-task self-supervised learning$^{\dagger}$ \cite{saeed2019multi} & Unsupervised & 75.41 & 83.30 & 80.20 & 45.37 \\ 
		Convolutional autoencoder$^{\dagger}$ \cite{haresamudram2019role} & Unsupervised & 79.58 & 82.50 & 80.26 & 48.82 \\ 
		Masked reconstruction$^{\dagger}$ \cite{haresamudram2020masked} & Unsupervised & 76.81 & 88.02 & \textbf{81.89} & 49.31 \\ \hline 
		CPC (1D Conv Encoder) & Unsupervised & \textbf{80.97} & \textbf{\textcolor{green}{89.05}} & 81.65 & \textbf{52.01} \\ \hline
	\end{tabular}
	\label{tab:representation_learning}
\end{table}

We perform CPC-based self-supervised pre-training and integrate the learned weights as a feature extractor in the activity recognition chain. 
In order to evaluate these learned representations, we compute their performance on the classifier network (Sec.\ \ref{sec:backend_classifier}).
The performance obtained by CPC is contrasted primarily against previous unsupervised approaches including multi-task self-supervised learning \cite{saeed2019multi}, convolutional autoencoders \cite{haresamudram2019role}, and masked reconstruction \cite{haresamudram2020masked}.
For reference, we also compare the performance relative to the supervised baseline--DeepConvLSTM \cite{ordonez2016deep}-- and a network with the same architecture as CPC, albeit trained end-to-end from scratch.
Once the model was pre-trained using CPC, the learned weights (from both $g_{enc}$ and $g_{ar}$) were frozen and used with the classifier network.
Labeled data was utilized to train the classifier network using cross entropy loss and the test set mean F1-score was detailed in Tab.\ \ref{tab:representation_learning}.

We first compare the performance of the CPC-based pre-training to state-of-the-art unsupervised learning approaches.
We note that all unsupervised learning approaches are evaluated on the same classifier network (Sec.\ \ref{sec:backend_classifier}), which is optimized during model training for activity recognition.
On Mobiact, Motionsense and USC-HAD, CPC-based pre-training outperforms \textbf{all} state-of-the-art unsupervised approaches.
For UCI-HAR, the performance is comparable to masked reconstruction.
This clearly demonstrates the effectiveness of the pre-training thereby fulfilling one of the goals of the paper -- which is to develop effective unsupervised pre-training approaches. 
It also validates our hypothesis that explicitly incorporating temporality at the representation level itself is beneficial towards learning useful representations.

Relative to DeepConvLSTM, i.e., the supervised baseline, the proposed approach shows increased performance by approximately $4\%$ on Motionsense and over $7\%$ on USC-HAD.
On UCI-HAR and Mobiact, the performance is comparable to the 1D Conv Encoder-based CPC model (last row). 
Relative to end-to-end training on a network with the same architecture as CPC, using the pre-training results in improved performance for Motionsense, which obtains the highest F1-score across both supervised and unsupervised learning approaches. 
In contrast, CPC-based pre-training performs comparably on UCI-HAR, while it shows reduced results for both USC-HAD and Mobiact  with the difference being around $3\%$ at worst.
This strong performance showcases the quality of the weights learned as they outperform end-to-end training on some of the datasets, even  when the number of trainable parameters are only a fraction (as only the final classifier parameters are updated).

\subsection{Semi-Supervised Learning on Limited Labeled Data}
\label{sec:semi_sup}

\begin{figure*}[t]
	\centering
	\includegraphics[width=0.9\textwidth]{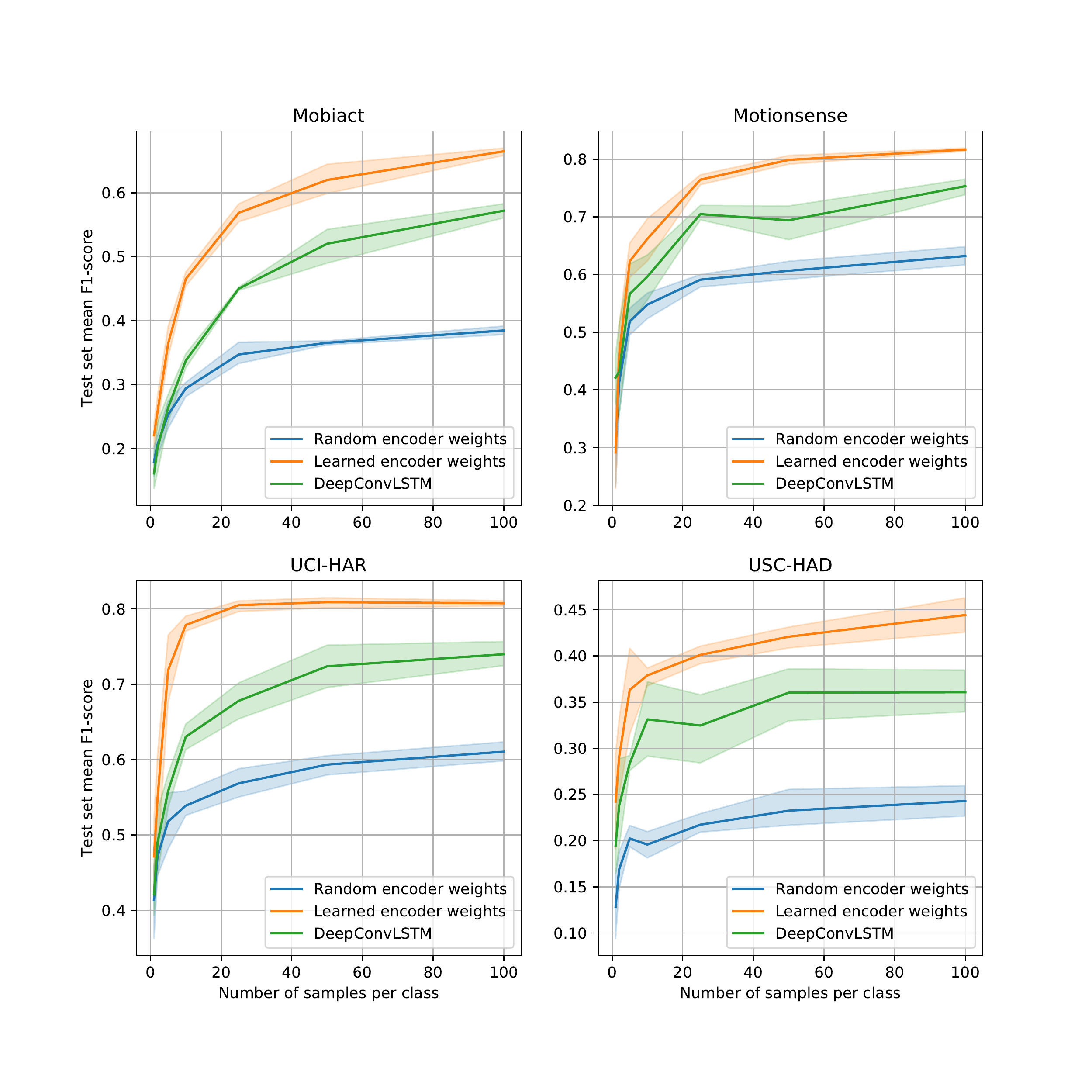}
	\vspace{-1 cm}
	\caption{Semi-supervised learning on limited labeled data. 
		The network is first pre-trained in an unsupervised manner using CPC, and the learned weights are frozen.
		The backend classifier is initialized randomly and trained from scratch using $\{$1, 2, 5, 10, 25, 50, 100$\}$ randomly sampled labelled frames.
		We perform five runs and report the test set mean f1-score.
		The curve in orange corresponds to CPC where the weights are learned while the blue line refers to the scenario with the randomly initialized feature extractor.
		The supervised baseline, DeepConvLSTM, is depicted in green.
		We observe significant improvements over DeepConvLSTM on all datasets.
	}
	\label{fig:limited_data}
\end{figure*}

Here we consider a very important scenario where only a small portion of the collected data is labeled. 
This is a practical and rather common situation in mobile and ubiquitous computing scenarios as real-world deployment may allow for acquiring only a small labeled dataset from the users with minimal interruptions. 
This scenario is important because data collection is typically straightforward yet annotation is often challenging. 

As in Sec.\ \ref{sec:rep_learning}, we first pre-train the network using the CPC task and utilize the weights learned on both the encoder and autoregressive network for feature extraction.
The classifier network is initialized randomly and trained from scratch whereas the learned weights are frozen. 
%
For each dataset, we utilize the entire dataset (unlabeled) for pre-training. 
During the fine-tuning stage, we randomly sample $x$ labeled samples per class where $x \in \{1, 2, 5, 10, 25, 50, 100\}$ and train the classifier from scratch.
We perform five runs and plot the test set mean F1-score in Fig.\ \ref{fig:limited_data} for the 1D Conv encoder models detailed in Tab.\ \ref{tab:representation_learning}.
The performance of the CPC-based pre-training technique is contrasted against the DeepConvLSTM network, which is the supervised learning baseline.
In order to make the contribution of the learned weights clearer, we also compute the performance when the feature extractor is initialized randomly and frozen during fine-tuning.

On Mobiact, which contains the largest number of participants, we immediately notice the improvement in performance even when only one labeled sample per class is available for fine-tuning. 
Relative to DeepConvLSTM, the performance of the proposed method is consistently higher by over $5\%$ throughout. 
The difference is even clearer over the random initialized feature extractor, as the pre-training results in an improvement of over $15\%$ when more than ten samples are available per class. 

In the case of Motionsense, DeepConvLSTM outperforms CPC when there is only one labeled samples per class. 
Beyond that, CPC shows increased performance, resulting in a maximum improvement of approx.\ $10\%$ when 50 labeled samples are available.
The learned weights result in sustained improvements of around $15\%$ over the random weights when more than ten labeled samples per class are available.

For UCI-HAR, CPC results show increased performance over DeepConvLSTM even when just one labeled sample per class is available. 
The difference peaks at over $10\%$ when we have access to ten samples per class.
Similarly, CPC shows improved performance even when there is only one sample per class for USC-HAD. 
As more labeled data is available, the performance of CPC rises further over DeepConvLSTM. 
There is also a consistent difference in performance over using a randomly initialized feature extractor.

The consistently improved performance over end-to-end training with DeepConvLSTM makes a compelling indicator that incorporating temporal characteristics into representation learning results in improved representations.
We also note that the largest unlabeled dataset, Mobiact, demonstrates the most significant improvement in performance over random weights due to the self-supervised pre-training.
Thus, using an even larger unlabeled dataset might further improve the quality of the representations learned using self-supervision.
Additionally, the confidence interval for the proposed technique, CPC, is generally narrower than DeepConvLSTM for all datasets.
This indicates that the pre-training makes the activity recognition performance more robust to variability in the input windows.

%

\section{Discussion}
The main hypothesis of our work is that explicitly targeting the temporal characteristics of movement data for representation learning is beneficial towards learning discriminative features for human activity recognition using body-worn sensors.
We accomplish this by predicting multiple future timesteps using contrastive learning. 
Such a pre-training scheme fits directly into the traditional activity recognition chains at the feature extraction step, allowing for easy integration into activity recognition workflows.
In what follows, we first analyze contrastive predictive coding as it is applied to HAR and derive insights on the performance of the pre-training. 
Then we look at the practical implications of our work on existing and future applications in mobile and ubiquitous computing and outline the research agenda related to representation learning for human activity recognition.

\subsection{Analyzing Contrastive Predictive Coding}
Contrastive predictive coding involves correctly identifying the positive sample from distractors for $k$ future timesteps.
In this section, we design experiments to better understand the pre-training process. 
We begin by studying the encoder ($g_{enc}$) and evaluate different choices, including feedforward, convolutional and recurrent networks and analyze how difficult the pretext task training is, and how it affects downstream activity recognition.
Subsequently, we also analyze the number of steps to predict during the pre-training.
Finally, we examine if all the weights learned during self-supervision are useful, by selectively utilizing only portions of the learned weights.
These experiments help us to build better self-supervised models.

\subsubsection{Choice of encoder}
\label{sec:choice_of_enc}

\begin{figure}[t]
	\centering
	\includegraphics[width=\textwidth]{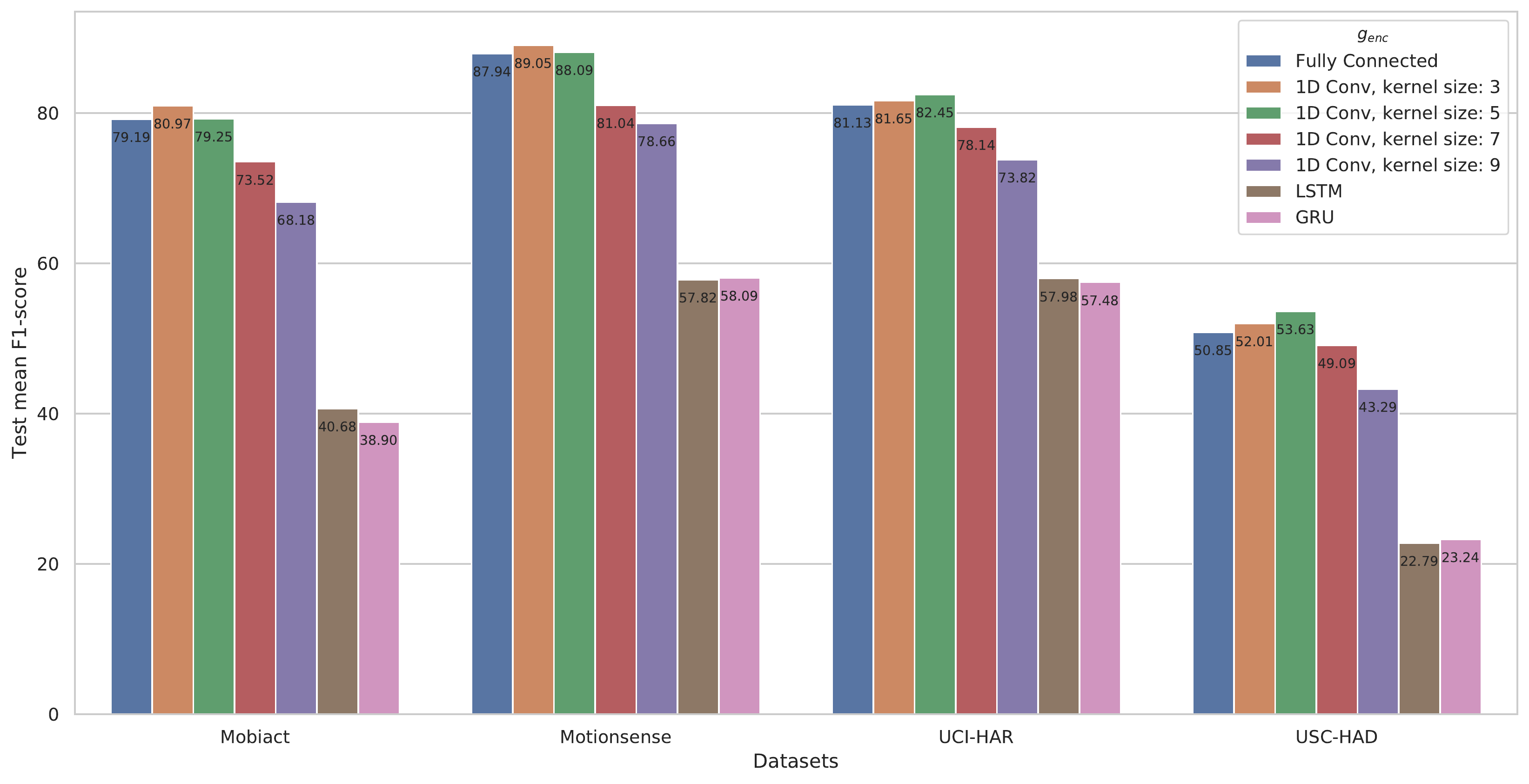}
	\caption{
		Studying the effect of using different $g_{enc}$ networks for CPC pre-training:  
		we utilize three categories of encoder networks including fully connected layers, 1D convolutional layers with kernel sizes $\in \{3, 5, 7, 9\}$, and recurrent layers  for pre-training.
		The learned $g_{enc}$ and $g_{ar}$ weights are frozen and used to extract representations for classifying activities. 
		We observe that the fully connected encoder or the 1D convolutional encoders with smaller kernel sizes (i.e. 3 or 5) demonstrate the highest activity recognition performance.
		}
	\label{fig:diff_encoders}
\end{figure}

\begin{figure}[t]
	\centering
	\includegraphics[width=\textwidth]{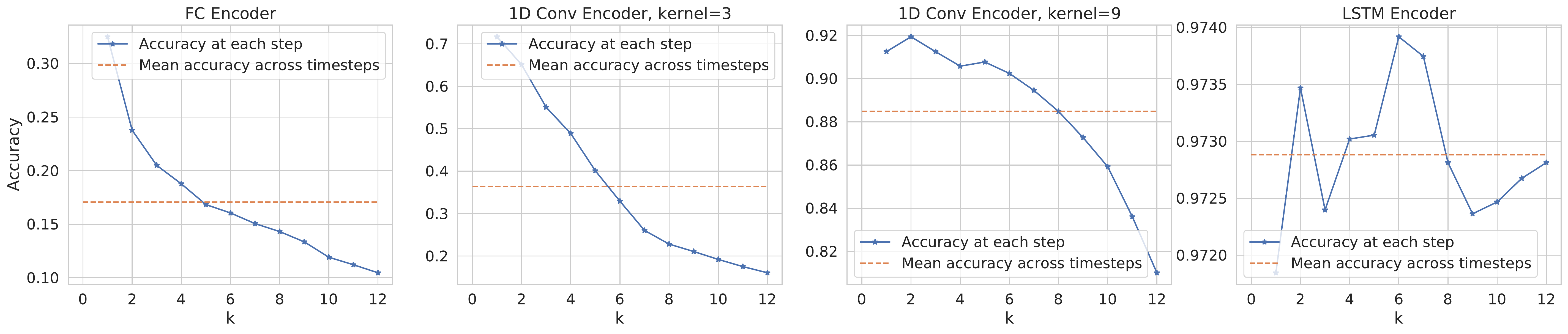}
	\caption{
		Accuracy of correctly predicting the positive sample from negatives in the contrastive loss, across multiple future timesteps ($k=12$) for test split of the Mobiact dataset.}
	\label{fig:all_accuracy_steps}
\end{figure}

The encoder $g_{enc}$ is used to map samples from the input data into a latent space on which future predictions are made. 
Therefore, the architecture of $g_{enc}$ has impact on the pre-training as well as the subsequent activity recognition. 
We study three categories of networks for $g_{enc}$ and report the downstream activity recognition performance in Fig.\ \ref{fig:diff_encoders}: 
\begin{description}
	\item[Fully Connected:] which consists of a feedforward network comprising of three fully connected layers with 32, 64 and 128 units, along with dropout~\cite{srivastava2014dropout} of p=$0.2$ and the ReLU activation function \cite{nair2010rectified} applied between consecutive fully connected layers.

	\item[1D Convolutional:] containing three 1D convolutional blocks with 32, 64 and 128 channels respectively. Each block consists of a 1D convolutional layer with reflect padding, followed by the ReLU activation and dropout with p=$0.2$. 
	We vary the kernel sizes $\in \{3, 5, 7, 9\}$.

	\item[Recurrent:] consisting of one layer long short-term memory (LSTM) or GRU network with 128 units. 
\end{description}
For each type of encoder, we utilize the same training settings as used in Tab.\ \ref{tab:representation_learning} for the 1D Conv Encoder.

Fig.\ \ref{fig:diff_encoders} shows that the choice of the encoder network has a significant impact on the activity recognition performance. 
For the 1D Conv. encoders, we see that having shorter kernel sizes (i.e., $3$ or $5$) is generally more preferable to longer filter lengths such as $7$ or $9$. 
The convolution operation over the input data results in overlaps over the future timesteps to be predicted.
Correctly predicting the first few future timesteps, which potentially fall under the overlap, thereby becomes trivial. 
This results in reduced activity recognition performance as the pretext task becomes ineffective. 
It also explains how the activity recognition performance reduces as the kernel size is increased.

Larger overlaps on the future timesteps mean that the overlapped steps are easier to predict, reducing the discriminative ability of the representations learned. 
This can be clearly seen in Fig.\ \ref{fig:all_accuracy_steps} where the 1D Conv. Encoder with a kernel size of $9$ has a considerably higher average accuracy of predicting the positive sample relative to using the encoder with a kernel size of $3$. 
It indicates that the pretext task is easier to solve.
Correspondingly, the downstream activity recognition performance for kernel size $9$ is around $13\%$ lower than the F1-score for kernel size $3$. 
This effect is more prominent for the recurrent encoders as the input data is parsed step-by-step and the output at any given timestep is dependent on previous data.
Thus, the pretext task becomes trivial to solve, resulting in a reduction of around $42\%$ compared to the 1D Conv. Encoder with kernel size $3$. 

The fully connected (FC) encoder represents the other extreme where there is no overlap across the window of data.
The average accuracy of predicting the future timesteps is lower relative to the 1D Conv. Encoder with kernel size $3$.
This results in reduction of nearly $2\%$ for activity recognition.
From Fig.\ \ref{fig:all_accuracy_steps} and Fig.\ \ref{fig:diff_encoders} we can see that the difficulty of solving the pretext task has significant impact on the activity recognition performance. 
If the task is trivial (as in the case of using large kernel sizes or recurrent encoders), the downstream recognition task also suffers correspondingly. 
Therefore, the encoder architecture must be carefully considered before being utilized for pre-training, as rendering the pretext task trivial results in poor representations.

\subsubsection{Number of Future Steps to Predict}
\begin{figure}[t]
	\centering
	\includegraphics[width=\textwidth]{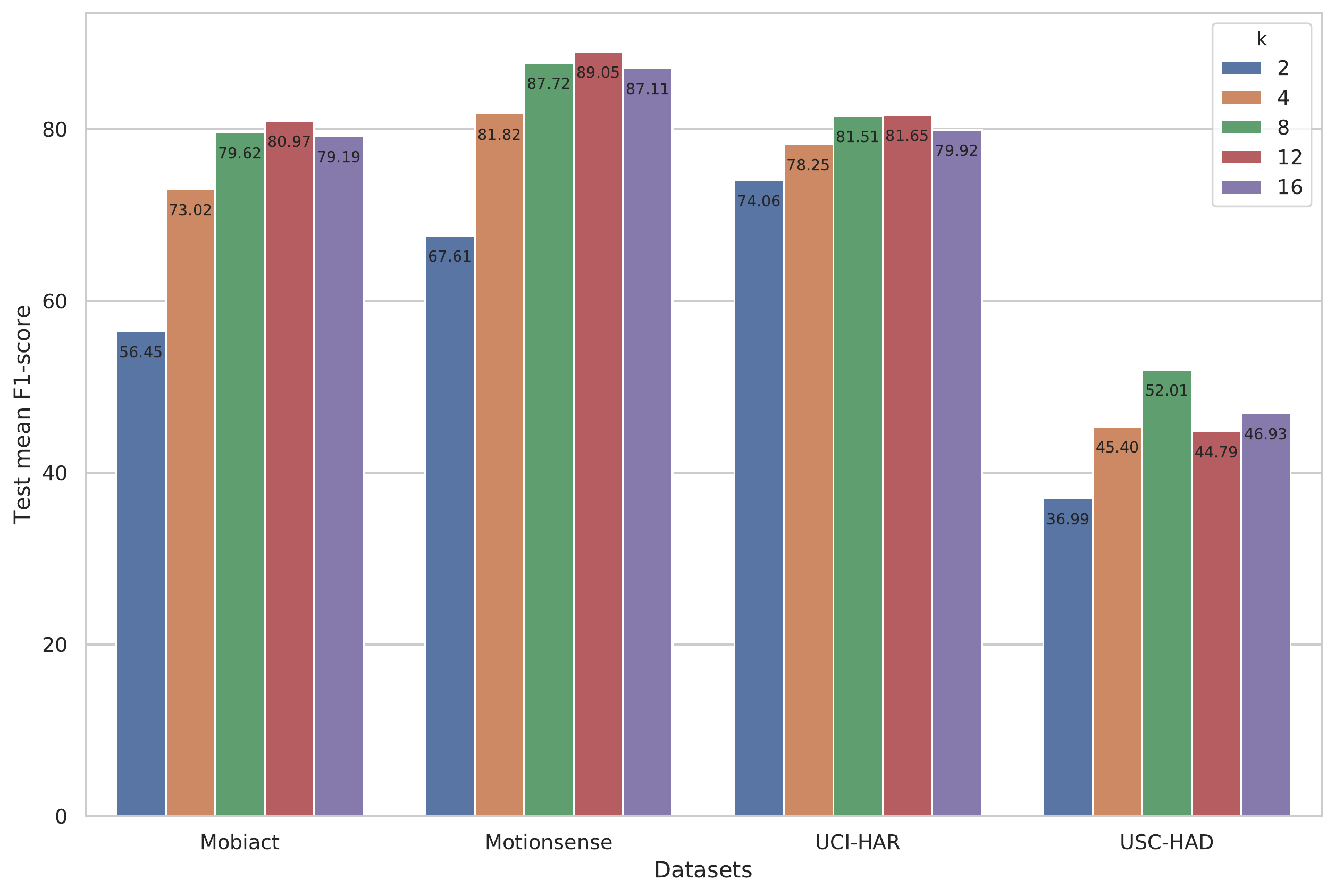}
	\caption{
		Studying the effect of the number of future timestep predictions (i.e. $k$) during CPC pre-training for activity recognition: 
		we utilize the 1D convolution encoder with a kernel size=3 as $g_{enc}$ and vary $k \in \{2, 4, 8, 12, 16\}$ for pre-training.
		Subsequently, the learned $g_{enc}$ and $g_{ar}$ weights are frozen and used as a feature extractor for activity recognition. 
		We observe that predicting multiple future timesteps ($k \geq 8$) is more advantageous than predicting  the near future, i.e. $k \in \{2, 4\}$. 
		}
	\label{fig:diff_ks}
\end{figure}

CPC learns effective representations by predicting multiple future timesteps, rather than just the next timestep.
This allows the model to learn features that incorporate long term temporal characteristics.
It also leads us to the question: \emph{How many future timesteps need to be predicted to learn useful representations?}
For the CPC-based pre-trained models in Tab.\ \ref{tab:representation_learning}, we study the activity recognition performance for increasing values of $k \in \{2, 4, 8, 12, 16\}$, and report the test mean F1-scores.

As can be seen in Fig.\ \ref{fig:diff_ks}, increasing the number of predicted timesteps generally results in improved activity recognition.
For Mobiact, Motionsense and UCI-HAR, the performance peaks when $k$=$12$. 
Beyond that, the performance starts reducing again. 
The trend is similar for USC-HAD, with the peak occurring at $k$=$8$.
Thus, predicting approximately 400ms into the future results in the best performance for Mobiact, Motionsense and UCI-HAR, whereas USC-HAD requires predicting around 270ms into the future.

As discussed in Sec.\ \ref{sec:cpc}, the input to the autoregressive network $g_{ar}$ is the sensor data between timesteps $0$ and $t$.
With increasing $k$, the input to $g_{ar}$ becomes shorter thereby making it more difficult to learn the context representation $c_t$ which can reliably predict many subsequent future timesteps.
We hypothesize that this results in the reduction in performance after obtaining the peak. 
Looking at the 1D Conv. Encoder with kernel size=3 in Fig.\ \ref{fig:all_accuracy_steps}, we can also observe that predicting multiple timesteps into the future is considerably more challenging than predicting the immediate future. 
The farther into the future we predict, the lower the mutual information between what the model already knows (i.e., the context representation) and the target, making it more difficult to correctly predict the positive sample. 
The accuracy of prediction drops almost exponentially indicating the difficulty of predicting far into the future.
However, Fig.\ \ref{fig:diff_ks} also shows that predicting the immediate future (i.e., $k=2$ in Fig.\ \ref{fig:diff_ks}) results in poorer representations as the model only learns short term noises. 
Therefore, predicting multiple future timesteps is vital towards extracting the long-term temporal signal which exists in the sensor data windows.

\subsubsection{Which Pre-Trained Weights Should be Used?}
\begin{figure}[t]
	\centering
	\includegraphics[width=\textwidth]{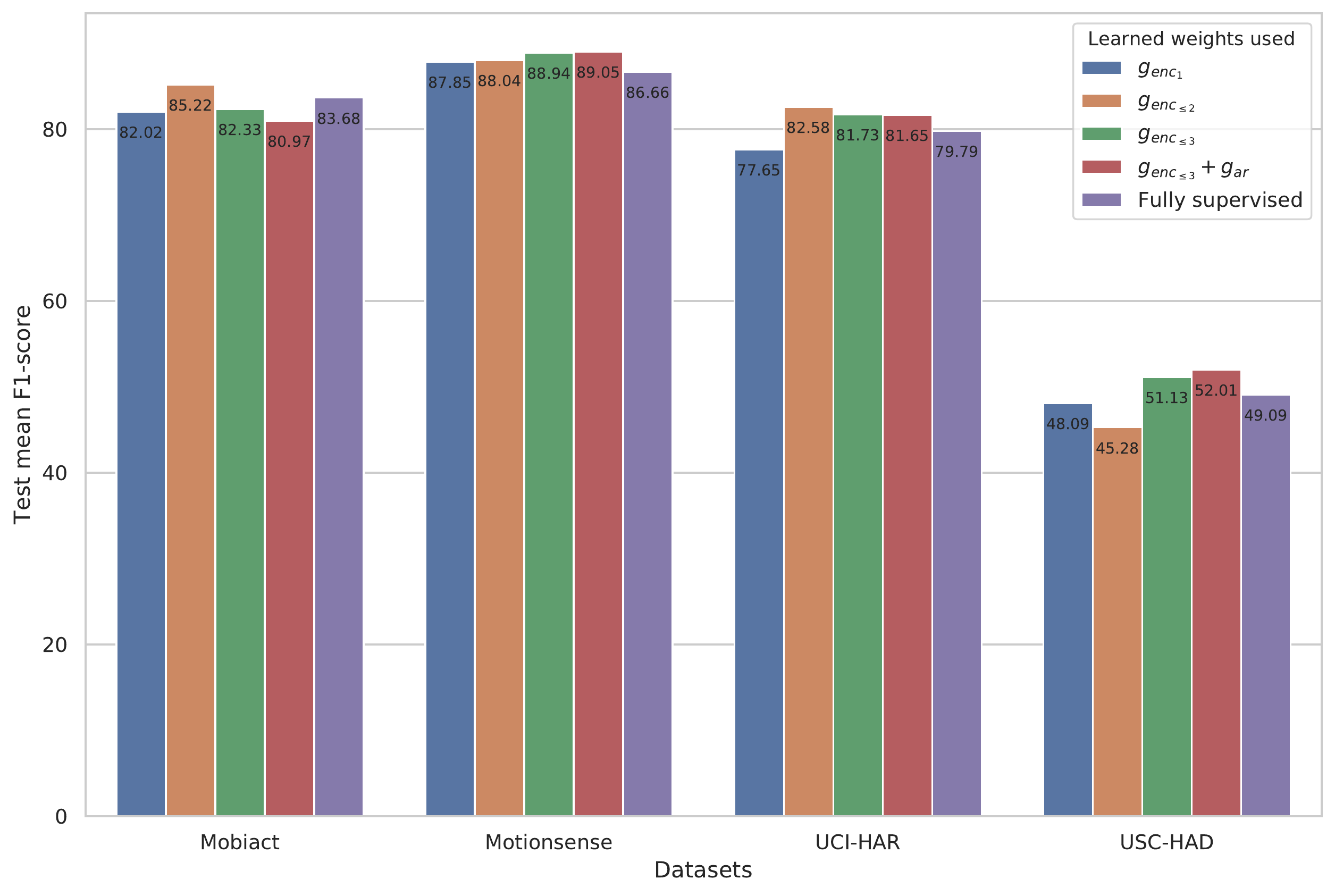}
	\caption{
		Studying which learned weights are useful for activity recognition: using learned weights from the CPC model with 1D convolutional encoder and kernel size=3, we progressively utilize learned weights from fewer layers for activity recognition.
		The rest of the layers are initialized randomly and optimized during activity recognition.
		For example, using both $g_{enc}$ and $g_{ar} $ weights is denoted by $g_{enc_{\leq3}} + g_{ar}$ while using only the first two encoder layer learned weights is shown as $g_{enc_{\leq2}}$ (the rest of the network including the third encoder layer and $g_{ar}$ are not frozen).
		We observe that using a subset of the learned weights can lead to improvement over using all learned weights. 		
	 }
	\label{fig:diff_encoder_levels}
\end{figure}

During activity recognition, the learned weights from both $g_{enc}$ and $g_{ar}$ were frozen and utilized with the classifier network (Fig.\ \ref{fig:cpc}).
This brings up the question: \emph{Which learned weights are useful for activity recognition?}
To answer this, we progressively utilize learned weights from fewer encoder layers during the classification and report the resulting test mean F1-score in Fig.\ \ref{fig:diff_encoder_levels}.

The default configuration involves freezing learned weights from both $g_{enc}$ and $g_{ar}$, as done for Tab.\ \ref{tab:representation_learning}.
This corresponds to $g_{enc_{\leq 3}} + g_{ar}$ in Fig.\  \ref{fig:diff_encoder_levels}.
$g_{enc_{\leq3}}$ utilizes frozen learned weights from all three encoder layers, while the autoregressive network $g_{ar}$ is initialized randomly and trained with the classifier.
On a similar vein, $g_{enc_{\leq2}}$ uses frozen learned weights from the first two encoder layers while the third layer and the autoregressive network $g_{ar}$ are both trained with the classifier.

From Fig.\ \ref{fig:diff_encoder_levels} we can see that utilizing the default configuration results in the best activity recognition performance for Motionsense and USC-HAD. 
For these datasets, using learned $g_{enc_{\leq3}} + g_{ar}$ weights outperforms fully supervised training (purple bars in Fig.\ \ref{fig:diff_encoder_levels}).

For Mobiact, the default configuration does not result in performance exceeding fully supervised training. 
However, using the learned $g_{enc_{\leq2}}$ weights results in a mean F1-score of $85.22\%$, which is an improvement over fully supervised training.
Similarly for UCI-HAR, the best performance is obtained when using the learned $g_{enc_{\leq2}}$ weights.
We observe a mean f1-score of $82.58\%$, which is higher than the best unsupervised representation learned by masked reconstruction (Tab.\ \ref{tab:representation_learning}) and comparable to DeepConvLSTM, which exceeds all other approaches.
Thus, we can conclude that using a subset of the weights learned via self-supervision could result in improved performance, often comparable to or better than supervised learning baselines such as DeepConvLSTM. 

\begin{figure}
	\centering
	\includegraphics[width=\textwidth]{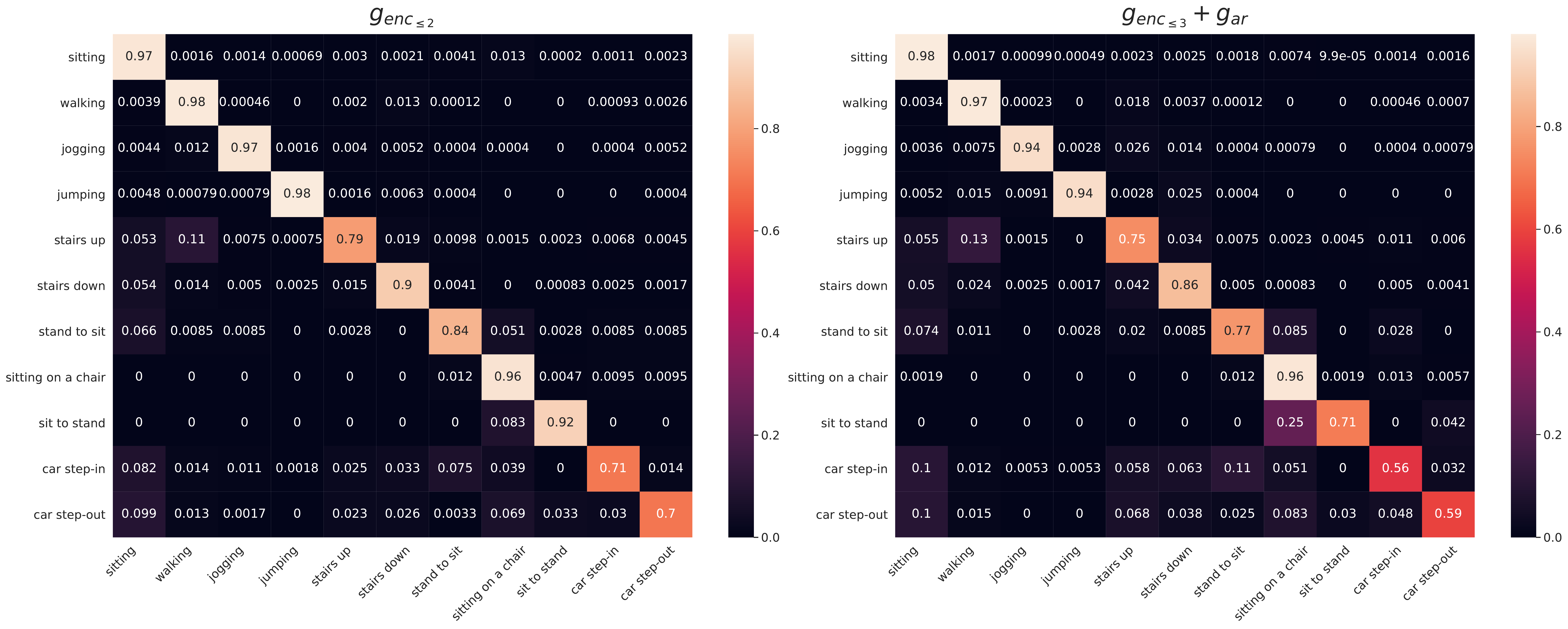}
	\caption{Confusion matrix for the activity recognition performed after using learned weights of different layers for the Mobiact dataset. 
	The pre-training was performed using 1D convolutional encoder with kernel size=3 and $k=12$.
	 $g_{enc_{\leq2}}$ (left) utilizes learned weights from only the first two convolutional layers whereas $g_{enc_{\leq 3}} + g_{ar}$ (right) uses learned weights from all encoder layers as well as $g_{ar}$.
	 For transition-style activities such as stand to sit, sit to stand, car step-in and car step-out, we see that the extra context provided by $g_{enc_{\leq2}}$ results in improved performance over $g_{enc_{\leq 3}} + g_{ar}$.
}
	\label{fig:confusionmatrix}
\end{figure}

As mentioned in \cite{oord2018representation}, either $c_t$ (the output of $g_{enc}$) or $z_t$ (the output of $g_{ar}$) can be utilized as the representation.
If extra context from the past is useful for classification, then $c_t$ can be used; else $z_t$ may be more effective.
We hypothesize that certain transition-style activities covered in Mobiact, such as stand to sit, sit to stand, car step-in and car step-out require the extra context provided by the learned weights at the encoder level, rather than the context representation ($c_t$), which may not be sufficiently encoding previous timesteps to be accurately classified. 
The confusion matrix in Fig.\ \ref{fig:confusionmatrix} illustrates this interpretation.
For the transition-style activities in particular, using the learned weights from only the first two encoder layers, i.e., $g_{enc_{\leq2}}$ results in considerable improvements over using $g_{enc_{\leq3}} + g_{ar}$. 

\subsubsection{Summary and Guidelines}
Studying factors which affect CPC such as the choice of encoder, number of future step predictions and the usability of learned weights at various levels of the network, allow us to develop a set of insights for applying CPC to other mobile and ubiquitous computing scenarios as well:
\begin{enumerate}
	\item The encoder architecture has significant impact on the quality of the representations learned via CPC. 
	Overlap over the prediction targets results in poorer representations, and thus, care must be taken to avoid setting up trivial prediction tasks.
	\item Predicting multiple future timesteps results in improved representations over predicting only the immediate future.
	Such a prediction scheme encodes global structure present in the time-series body-worn sensor data, which is beneficial towards learning richer representations.
	\item A subset of the learned weights could result in improved performance relative to using all learned weights.
	The representation used depends on the activities under study, as certain activities might require longer context to be accurately recognized. 
	Thus, using learned weights from both the encoder and autoregressive networks maybe sub-optimal than using some parts of those weights. 
\end{enumerate}

\subsection{Practical Value of CPC for Human Activity Recognition Tasks}
\subsubsection{Incorporating Temporality at Representation Level results in Improved Recognition Performance}
The primary hypothesis of this work is that incorporating temporal characteristics into the representation learning process results in effective representation learning for time-series data. 
We note that previous unsupervised learning approaches involving Restricted Boltzmann Machines\cite{plotz2011feature} and convolutional autoencoders \cite{haresamudram2019role} were trained using signal reconstruction.
The more recent multi-task self-supervised learning utilizes binary predictions of randomly applying binary signal transformations in order to learn representations. 
These approaches however, do not specifically utilize the temporal characteristics of the data. 
Masked reconstruction \cite{haresamudram2020masked} leverages local temporal dependencies by reconstructing zeroed out timesteps from immediate local context. 
From Tab.\ \ref{tab:representation_learning} we observe that masked reconstruction, using just local low-level information, results in improvements in performance on three out of four of the benchmark datasets over the approaches that do not seek to incorporate any level of temporal information.
This demonstrates the importance of using temporal information into the representation learning process itself. 

Taking this insight further, we proposed to utilize Contrastive Predictive Coding, which learns to infer global structure in the time-series data, beyond just local low-level information and noise, by predicting multiple future timesteps.
From our results in Tab.\ \ref{tab:representation_learning} we obtain strong evidence that such a process results in rich representations that more effectively capture the time-series nature of the body-worn sensor data. 

\subsubsection{CPC is Flexible and Generic}
The CPC framework can be considered generic from three standpoints:
\begin{description}
	\item[Target prediction setup:] The temporal order of prediction is not significant and the pretext task can also be setup to predict the past timesteps or even randomly masked portions of the data (similar to masked reconstruction).
	However, care needs to be taken while choosing the encoder, the target predictions as well as the negative samples in order to avoid setting up a trivial task.

	\item[Architecture:] The architecture of CPC also allows for replacing the $g_{enc}$, $g_{ar}$, and $W_k$ networks with suitable alternatives. 
	In this work we utilized standard architectures for simplicity. 
	For example, in Sec.\ \ref{sec:choice_of_enc} we discussed various encoder networks and their effects.
	The autoregressive network $g_{ar}$ used in this work is a 2-layer GRU. 
	However, more powerful techniques for processing sequential information such as Transformers \cite{vaswani2017attention} or convolution layers could further improve the representation performance. 
	Finally. the future timestep predictions are made by $W_k$, which in our setup consists of linear layers.
	They only observe the context representation in order to make the prediction.
	As explored in \cite{riviere2020unsupervised}, alternates to this setup include 1-layer Transformer encoders and recurrent networks.

	\item[How negatives are sampled:] 
	In this work, the negatives for the contrastive loss are sampled from different windows present in the same batch of data. 
	However, if we had prior knowledge of the downstream task, the negative sampling could be performed in such a way to incentivize relevant features for that particular task.
	For example, if the downstream task was person identification, we could sample negatives from the rest of the participants for each positive sample.
	The resulting features are likely to be discriminative towards participants rather than activities. 
\end{description}

\subsubsection{Self-Supervised Pre-Training Reduces Reliance on Large Amounts of Annotated Data}
The lack of large-scale annotated body-worn sensor datasets for human activity recognition limits the application of complex supervised recognition systems. 
At the same time, the more straightforward data collection process allows for the potential collection of large amounts of unlabeled data. 
In this work, we demonstrated how pre-training on unlabeled data results in performance exceeding fully supervised training when limited labeled data is available. 
This allows for leveraging massive movement datasets such as the UK Biobank \cite{doherty2017large} to learn general movement representations, that can subsequently be fine-tuned for target tasks using labeled data.
Such an approach can have serious impact in practical scenarios where very limited data can be collected from users after the deployment of recognition systems, for user adaptation or personalization purposes. 
By first pre-training on unlabeled data, these limited labels can be economically used to derive accurate recognition.

\subsection{Research Agenda: Representation Learning for Human Activity Recognition}
The focus of this work is towards utilizing unlabeled movement sensor datasets to perform representation learning.
It is motivated by the limited size and the lack of variety in current annotated datasets.
Instead, we can leverage massive movement datasets such as the UK Biobank \cite{doherty2017large} to first learn generic representations that can be fine-tuned on task specific labeled data. 
However, specific applications of mobile and ubiquitous computing might require sensor placements which are different than the available source dataset.
Therefore, a big challenge to address is the development of approaches that can still learn representations that can be effectively fine-tuned on data collected from sensors placed on different locations on the body.
This will allow for improved performance in niche scenarios requiring unique sensor placements.

Recent approaches addressing the lack of annotated datasets involve extraction of virtual movement data from modalities such as mocap \cite{huang2018deep, takeda2018multi} and videos \cite{kwon2020imutube}, which contain a large number of participants and diverse set of actions and environments.
Such techniques have the advantage of being capable of extracting large quantities of virtual movement data at arbitrary positions on the body.
However, due to the artifacts introduced during the extraction process, the performance of recognizers trained on this data tend to be lower than those trained on real IMU data. 
Thus, developing effective representation learning approaches that can deal with the domain shift could be beneficial towards improved recognition.

\section{Conclusion}
Feature extraction plays a crucial role in human activity recognition (HAR) using body-worn movement sensors.
Accordingly, the community has invested a lot of effort into developing new ways to extract meaningful representations from streams of sensor data that, ultimately, will lead to improved downstream recognition performance.
Hand-crafted (engineered) features have long dominated the field, yet with the success of deep learning methods in the wider scientific community, the focus of the HAR community has shifted to learning effective representations directly from sensor data.
End-to-end learning can only be employed with some limitations given the typically severe restrictions on the amount of \textit{labelled} sample data that is available for training HAR models from scratch.
As such, substantial effort has been devoted to explicit representation learning thereby not necessarily relying on large quantities of labeled sample data but rather making more economic and hence effective use of smaller labeled training sets, and especially aiming at exploiting unlabeled data, which are straightforward to collect in mobile and ubiquitous computing settings.

The approach presented in this paper follows the paradigm of explicitly learning data representations, which are then integrated as features into the standard activity recognition chain \cite{bulling2014tutorial}.
The key idea for the presented approach is to focus directly on temporality in the data in order to learn feature representations that lead to improved activity recognition performance especially for challenging scenarios with limited labeled training data.
To this end, we have introduced the concept of Contrastive Predictive Coding (CPC) into human activity recognition using body-worn movement sensors. 
CPC was demonstrated to be effective in, for example, audio analysis scenarios and we have adopted the technique and refined it towards the constraints and requirements of HAR.
CPC learns features through a combination of encoder networks that not only target representing individual sample but rather focus on predicting samples in the temporal vicinity and as such explicitly aiming at modeling temporality.
We have hypothesized that this temporality is crucial not only at modeling level but especially at representation level.
The pre-trained models are then integrated into the activity recognition chain, serving as effective feature extractors.

In our extensive experimental evaluation we have demonstrated the effectiveness of the proposed approach for improved human activity recognition under realistic, challenging requirements.
On a range of benchmark scenarios we have shown that CPC-learned features lead to recognition models that outperform all previous approaches to unsupervised representation learning.
Furthermore, we have demonstrated that the CPC-based models are on par with supervised learning approaches. 
Yet, such end-to-end methods are not suitable for real-world application scenarios with limited annotated training sets.
We have shown how CPC-based models can overcome these issues by demonstrating that recognition results are significantly better for small, labeled sampled sets when compared to the state-of-the-art in end-to-end learning.

These results are encouraging because they indicate that it is possible to utilize unlabeled data for deriving effective sensor data representations that in turn will lead to more effective recognition systems.
Collecting even vast amounts of unlabelled sensor data can be considered straightforward given the ubiquity of mobile sensing platforms such as smartphones or other body-worn sensors. 
The work presented here adds to the general agenda of deriving robust and effective recognition systems for challenging assessment scenarios as they are common in applications of mobile and ubiquitous computing.



\bibliographystyle{ACM-Reference-Format}
\bibliography{refs}


\begin{thebibliography}{88}


\ifx \showCODEN    \undefined \def \showCODEN     #1{\unskip}     \fi
\ifx \showDOI      \undefined \def \showDOI       #1{#1}\fi
\ifx \showISBNx    \undefined \def \showISBNx     #1{\unskip}     \fi
\ifx \showISBNxiii \undefined \def \showISBNxiii  #1{\unskip}     \fi
\ifx \showISSN     \undefined \def \showISSN      #1{\unskip}     \fi
\ifx \showLCCN     \undefined \def \showLCCN      #1{\unskip}     \fi
\ifx \shownote     \undefined \def \shownote      #1{#1}          \fi
\ifx \showarticletitle \undefined \def \showarticletitle #1{#1}   \fi
\ifx \showURL      \undefined \def \showURL       {\relax}        \fi
\providecommand\bibfield[2]{#2}
\providecommand\bibinfo[2]{#2}
\providecommand\natexlab[1]{#1}
\providecommand\showeprint[2][]{arXiv:#2}

\bibitem[\protect\citeauthoryear{Abowd}{Abowd}{2012}]%
        {Abowd:2012vt}
\bibfield{author}{\bibinfo{person}{Gregory~D Abowd}.}
  \bibinfo{year}{2012}\natexlab{}.
\newblock \showarticletitle{{What next, Ubicomp? Celebrating an intellectual
  disappearing act}}.
\newblock \bibinfo{journal}{\emph{Proc. Int. Conf. on Ubiquitous Computing
  (UbiComp)}} (\bibinfo{year}{2012}).
\newblock


\bibitem[\protect\citeauthoryear{Amft, Junker, and Troster}{Amft
  et~al\mbox{.}}{2005}]%
        {amft2005detection}
\bibfield{author}{\bibinfo{person}{Oliver Amft}, \bibinfo{person}{Holger
  Junker}, {and} \bibinfo{person}{Gerhard Troster}.}
  \bibinfo{year}{2005}\natexlab{}.
\newblock \showarticletitle{Detection of eating and drinking arm gestures using
  inertial body-worn sensors}. In \bibinfo{booktitle}{\emph{Ninth IEEE
  International Symposium on Wearable Computers (ISWC'05)}}. IEEE,
  \bibinfo{pages}{160--163}.
\newblock


\bibitem[\protect\citeauthoryear{Amodei, Ananthanarayanan, Anubhai, Bai,
  Battenberg, Case, Casper, Catanzaro, Cheng, Chen, et~al\mbox{.}}{Amodei
  et~al\mbox{.}}{2016}]%
        {amodei2016deep}
\bibfield{author}{\bibinfo{person}{Dario Amodei}, \bibinfo{person}{Sundaram
  Ananthanarayanan}, \bibinfo{person}{Rishita Anubhai},
  \bibinfo{person}{Jingliang Bai}, \bibinfo{person}{Eric Battenberg},
  \bibinfo{person}{Carl Case}, \bibinfo{person}{Jared Casper},
  \bibinfo{person}{Bryan Catanzaro}, \bibinfo{person}{Qiang Cheng},
  \bibinfo{person}{Guoliang Chen}, {et~al\mbox{.}}}
  \bibinfo{year}{2016}\natexlab{}.
\newblock \showarticletitle{Deep speech 2: End-to-end speech recognition in
  english and mandarin}. In \bibinfo{booktitle}{\emph{International conference
  on machine learning}}. \bibinfo{pages}{173--182}.
\newblock


\bibitem[\protect\citeauthoryear{Anguita, Ghio, Oneto, Parra, and
  Reyes-Ortiz}{Anguita et~al\mbox{.}}{2013}]%
        {anguita2013public}
\bibfield{author}{\bibinfo{person}{Davide Anguita}, \bibinfo{person}{Alessandro
  Ghio}, \bibinfo{person}{Luca Oneto}, \bibinfo{person}{Xavier Parra}, {and}
  \bibinfo{person}{Jorge~Luis Reyes-Ortiz}.} \bibinfo{year}{2013}\natexlab{}.
\newblock \showarticletitle{A public domain dataset for human activity
  recognition using smartphones.}. In \bibinfo{booktitle}{\emph{Esann}}.
\newblock


\bibitem[\protect\citeauthoryear{Atal and Schroeder}{Atal and
  Schroeder}{1970}]%
        {atal1970adaptive}
\bibfield{author}{\bibinfo{person}{Bishnu~S Atal} {and}
  \bibinfo{person}{Manfred~R Schroeder}.} \bibinfo{year}{1970}\natexlab{}.
\newblock \showarticletitle{Adaptive predictive coding of speech signals}.
\newblock \bibinfo{journal}{\emph{Bell System Technical Journal}}
  \bibinfo{volume}{49}, \bibinfo{number}{8} (\bibinfo{year}{1970}),
  \bibinfo{pages}{1973--1986}.
\newblock


\bibitem[\protect\citeauthoryear{Baevski, Zhou, Mohamed, and Auli}{Baevski
  et~al\mbox{.}}{2020}]%
        {baevski2020wav2vec}
\bibfield{author}{\bibinfo{person}{Alexei Baevski}, \bibinfo{person}{Henry
  Zhou}, \bibinfo{person}{Abdelrahman Mohamed}, {and} \bibinfo{person}{Michael
  Auli}.} \bibinfo{year}{2020}\natexlab{}.
\newblock \showarticletitle{wav2vec 2.0: A framework for self-supervised
  learning of speech representations}.
\newblock \bibinfo{journal}{\emph{arXiv preprint arXiv:2006.11477}}
  (\bibinfo{year}{2020}).
\newblock


\bibitem[\protect\citeauthoryear{Bedri, Li, Haynes, Kosaraju, Grover, Prioleau,
  Beh, Goel, Starner, and Abowd}{Bedri et~al\mbox{.}}{2017}]%
        {bedri2017earbit}
\bibfield{author}{\bibinfo{person}{Abdelkareem Bedri}, \bibinfo{person}{Richard
  Li}, \bibinfo{person}{Malcolm Haynes}, \bibinfo{person}{Raj~Prateek
  Kosaraju}, \bibinfo{person}{Ishaan Grover}, \bibinfo{person}{Temiloluwa
  Prioleau}, \bibinfo{person}{Min~Yan Beh}, \bibinfo{person}{Mayank Goel},
  \bibinfo{person}{Thad Starner}, {and} \bibinfo{person}{Gregory Abowd}.}
  \bibinfo{year}{2017}\natexlab{}.
\newblock \showarticletitle{EarBit: using wearable sensors to detect eating
  episodes in unconstrained environments}.
\newblock \bibinfo{journal}{\emph{Proceedings of the ACM on interactive,
  mobile, wearable and ubiquitous technologies}} \bibinfo{volume}{1},
  \bibinfo{number}{3} (\bibinfo{year}{2017}), \bibinfo{pages}{1--20}.
\newblock


\bibitem[\protect\citeauthoryear{Bulling, Blanke, and Schiele}{Bulling
  et~al\mbox{.}}{2014}]%
        {bulling2014tutorial}
\bibfield{author}{\bibinfo{person}{Andreas Bulling}, \bibinfo{person}{Ulf
  Blanke}, {and} \bibinfo{person}{Bernt Schiele}.}
  \bibinfo{year}{2014}\natexlab{}.
\newblock \showarticletitle{A tutorial on human activity recognition using
  body-worn inertial sensors}.
\newblock \bibinfo{journal}{\emph{ACM Computing Surveys (CSUR)}}
  \bibinfo{volume}{46}, \bibinfo{number}{3} (\bibinfo{year}{2014}),
  \bibinfo{pages}{1--33}.
\newblock


\bibitem[\protect\citeauthoryear{Cer, Yang, Kong, Hua, Limtiaco, John,
  Constant, Guajardo-Cespedes, Yuan, Tar, et~al\mbox{.}}{Cer
  et~al\mbox{.}}{2018}]%
        {cer2018universal}
\bibfield{author}{\bibinfo{person}{Daniel Cer}, \bibinfo{person}{Yinfei Yang},
  \bibinfo{person}{Sheng-yi Kong}, \bibinfo{person}{Nan Hua},
  \bibinfo{person}{Nicole Limtiaco}, \bibinfo{person}{Rhomni~St John},
  \bibinfo{person}{Noah Constant}, \bibinfo{person}{Mario Guajardo-Cespedes},
  \bibinfo{person}{Steve Yuan}, \bibinfo{person}{Chris Tar}, {et~al\mbox{.}}}
  \bibinfo{year}{2018}\natexlab{}.
\newblock \showarticletitle{Universal sentence encoder}.
\newblock \bibinfo{journal}{\emph{arXiv preprint arXiv:1803.11175}}
  (\bibinfo{year}{2018}).
\newblock


\bibitem[\protect\citeauthoryear{Chatzaki, Pediaditis, Vavoulas, and
  Tsiknakis}{Chatzaki et~al\mbox{.}}{2016}]%
        {chatzaki2016human}
\bibfield{author}{\bibinfo{person}{Charikleia Chatzaki},
  \bibinfo{person}{Matthew Pediaditis}, \bibinfo{person}{George Vavoulas},
  {and} \bibinfo{person}{Manolis Tsiknakis}.} \bibinfo{year}{2016}\natexlab{}.
\newblock \showarticletitle{Human daily activity and fall recognition using a
  smartphone’s acceleration sensor}. In
  \bibinfo{booktitle}{\emph{International Conference on Information and
  Communication Technologies for Ageing Well and e-Health}}. Springer,
  \bibinfo{pages}{100--118}.
\newblock


\bibitem[\protect\citeauthoryear{Chavarriaga, Sagha, Calatroni, Digumarti,
  Tr{\"o}ster, Mill{\'a}n, and Roggen}{Chavarriaga et~al\mbox{.}}{2013}]%
        {chavarriaga2013opportunity}
\bibfield{author}{\bibinfo{person}{R. Chavarriaga}, \bibinfo{person}{H. Sagha},
  \bibinfo{person}{A. Calatroni}, \bibinfo{person}{S.~T. Digumarti},
  \bibinfo{person}{G. Tr{\"o}ster}, \bibinfo{person}{J.~R. Mill{\'a}n}, {and}
  \bibinfo{person}{D. Roggen}.} \bibinfo{year}{2013}\natexlab{}.
\newblock \showarticletitle{The Opportunity challenge: A benchmark database for
  on-body sensor-based activity recognition}.
\newblock \bibinfo{journal}{\emph{Pattern Recognition Letters}}
  \bibinfo{volume}{34}, \bibinfo{number}{15} (\bibinfo{year}{2013}),
  \bibinfo{pages}{2033--2042}.
\newblock


\bibitem[\protect\citeauthoryear{Chen, Kornblith, Norouzi, and Hinton}{Chen
  et~al\mbox{.}}{2020}]%
        {chen2020simple}
\bibfield{author}{\bibinfo{person}{Ting Chen}, \bibinfo{person}{Simon
  Kornblith}, \bibinfo{person}{Mohammad Norouzi}, {and}
  \bibinfo{person}{Geoffrey Hinton}.} \bibinfo{year}{2020}\natexlab{}.
\newblock \showarticletitle{A simple framework for contrastive learning of
  visual representations}.
\newblock \bibinfo{journal}{\emph{arXiv preprint arXiv:2002.05709}}
  (\bibinfo{year}{2020}).
\newblock


\bibitem[\protect\citeauthoryear{Chuang, Robinson, Lin, Torralba, and
  Jegelka}{Chuang et~al\mbox{.}}{2020}]%
        {chuang2020debiased}
\bibfield{author}{\bibinfo{person}{Ching-Yao Chuang}, \bibinfo{person}{Joshua
  Robinson}, \bibinfo{person}{Yen-Chen Lin}, \bibinfo{person}{Antonio
  Torralba}, {and} \bibinfo{person}{Stefanie Jegelka}.}
  \bibinfo{year}{2020}\natexlab{}.
\newblock \showarticletitle{Debiased contrastive learning}.
\newblock \bibinfo{journal}{\emph{Advances in Neural Information Processing
  Systems}}  \bibinfo{volume}{33} (\bibinfo{year}{2020}).
\newblock


\bibitem[\protect\citeauthoryear{Chung, Gulcehre, Cho, and Bengio}{Chung
  et~al\mbox{.}}{2014}]%
        {chung2014empirical}
\bibfield{author}{\bibinfo{person}{Junyoung Chung}, \bibinfo{person}{Caglar
  Gulcehre}, \bibinfo{person}{KyungHyun Cho}, {and} \bibinfo{person}{Yoshua
  Bengio}.} \bibinfo{year}{2014}\natexlab{}.
\newblock \showarticletitle{Empirical evaluation of gated recurrent neural
  networks on sequence modeling}.
\newblock \bibinfo{journal}{\emph{arXiv preprint arXiv:1412.3555}}
  (\bibinfo{year}{2014}).
\newblock


\bibitem[\protect\citeauthoryear{Chung and Glass}{Chung and Glass}{2020a}]%
        {chung2020generative}
\bibfield{author}{\bibinfo{person}{Yu-An Chung} {and} \bibinfo{person}{James
  Glass}.} \bibinfo{year}{2020}\natexlab{a}.
\newblock \showarticletitle{Generative pre-training for speech with
  autoregressive predictive coding}. In \bibinfo{booktitle}{\emph{ICASSP
  2020-2020 IEEE International Conference on Acoustics, Speech and Signal
  Processing (ICASSP)}}. IEEE, \bibinfo{pages}{3497--3501}.
\newblock


\bibitem[\protect\citeauthoryear{Chung and Glass}{Chung and Glass}{2020b}]%
        {chung2020improved}
\bibfield{author}{\bibinfo{person}{Yu-An Chung} {and} \bibinfo{person}{James
  Glass}.} \bibinfo{year}{2020}\natexlab{b}.
\newblock \showarticletitle{Improved speech representations with multi-target
  autoregressive predictive coding}.
\newblock \bibinfo{journal}{\emph{arXiv preprint arXiv:2004.05274}}
  (\bibinfo{year}{2020}).
\newblock


\bibitem[\protect\citeauthoryear{Chung, Hsu, Tang, and Glass}{Chung
  et~al\mbox{.}}{2019}]%
        {chung2019unsupervised}
\bibfield{author}{\bibinfo{person}{Yu-An Chung}, \bibinfo{person}{Wei-Ning
  Hsu}, \bibinfo{person}{Hao Tang}, {and} \bibinfo{person}{James Glass}.}
  \bibinfo{year}{2019}\natexlab{}.
\newblock \showarticletitle{An unsupervised autoregressive model for speech
  representation learning}.
\newblock \bibinfo{journal}{\emph{arXiv preprint arXiv:1904.03240}}
  (\bibinfo{year}{2019}).
\newblock


\bibitem[\protect\citeauthoryear{Devlin, Chang, Lee, and Toutanova}{Devlin
  et~al\mbox{.}}{2018}]%
        {devlin2018bert}
\bibfield{author}{\bibinfo{person}{Jacob Devlin}, \bibinfo{person}{Ming-Wei
  Chang}, \bibinfo{person}{Kenton Lee}, {and} \bibinfo{person}{Kristina
  Toutanova}.} \bibinfo{year}{2018}\natexlab{}.
\newblock \showarticletitle{Bert: Pre-training of deep bidirectional
  transformers for language understanding}.
\newblock \bibinfo{journal}{\emph{arXiv preprint arXiv:1810.04805}}
  (\bibinfo{year}{2018}).
\newblock


\bibitem[\protect\citeauthoryear{Doersch, Gupta, and Efros}{Doersch
  et~al\mbox{.}}{2015}]%
        {doersch2015unsupervised}
\bibfield{author}{\bibinfo{person}{Carl Doersch}, \bibinfo{person}{Abhinav
  Gupta}, {and} \bibinfo{person}{Alexei~A Efros}.}
  \bibinfo{year}{2015}\natexlab{}.
\newblock \showarticletitle{Unsupervised visual representation learning by
  context prediction}. In \bibinfo{booktitle}{\emph{Proceedings of the IEEE
  international conference on computer vision}}. \bibinfo{pages}{1422--1430}.
\newblock


\bibitem[\protect\citeauthoryear{Doherty, Jackson, Hammerla, Pl{\"o}tz,
  Olivier, Granat, White, Van~Hees, Trenell, Owen, et~al\mbox{.}}{Doherty
  et~al\mbox{.}}{2017}]%
        {doherty2017large}
\bibfield{author}{\bibinfo{person}{Aiden Doherty}, \bibinfo{person}{Dan
  Jackson}, \bibinfo{person}{Nils Hammerla}, \bibinfo{person}{Thomas
  Pl{\"o}tz}, \bibinfo{person}{Patrick Olivier}, \bibinfo{person}{Malcolm~H
  Granat}, \bibinfo{person}{Tom White}, \bibinfo{person}{Vincent~T Van~Hees},
  \bibinfo{person}{Michael~I Trenell}, \bibinfo{person}{Christoper~G Owen},
  {et~al\mbox{.}}} \bibinfo{year}{2017}\natexlab{}.
\newblock \showarticletitle{Large scale population assessment of physical
  activity using wrist worn accelerometers: The UK Biobank Study}.
\newblock \bibinfo{journal}{\emph{PloS one}} \bibinfo{volume}{12},
  \bibinfo{number}{2} (\bibinfo{year}{2017}), \bibinfo{pages}{e0169649}.
\newblock


\bibitem[\protect\citeauthoryear{Elias}{Elias}{1955}]%
        {elias1955predictive}
\bibfield{author}{\bibinfo{person}{Peter Elias}.}
  \bibinfo{year}{1955}\natexlab{}.
\newblock \showarticletitle{Predictive coding--I}.
\newblock \bibinfo{journal}{\emph{IRE Transactions on Information Theory}}
  \bibinfo{volume}{1}, \bibinfo{number}{1} (\bibinfo{year}{1955}),
  \bibinfo{pages}{16--24}.
\newblock


\bibitem[\protect\citeauthoryear{Fernando, Bilen, Gavves, and Gould}{Fernando
  et~al\mbox{.}}{2017}]%
        {fernando2017self}
\bibfield{author}{\bibinfo{person}{Basura Fernando}, \bibinfo{person}{Hakan
  Bilen}, \bibinfo{person}{Efstratios Gavves}, {and} \bibinfo{person}{Stephen
  Gould}.} \bibinfo{year}{2017}\natexlab{}.
\newblock \showarticletitle{Self-supervised video representation learning with
  odd-one-out networks}. In \bibinfo{booktitle}{\emph{Proceedings of the IEEE
  conference on computer vision and pattern recognition}}.
  \bibinfo{pages}{3636--3645}.
\newblock


\bibitem[\protect\citeauthoryear{Figo, Diniz, Ferreira, and Cardoso}{Figo
  et~al\mbox{.}}{2010}]%
        {figo2010preprocessing}
\bibfield{author}{\bibinfo{person}{Davide Figo}, \bibinfo{person}{Pedro~C
  Diniz}, \bibinfo{person}{Diogo~R Ferreira}, {and} \bibinfo{person}{Joao~MP
  Cardoso}.} \bibinfo{year}{2010}\natexlab{}.
\newblock \showarticletitle{Preprocessing techniques for context recognition
  from accelerometer data}.
\newblock \bibinfo{journal}{\emph{Personal and Ubiquitous Computing}}
  \bibinfo{volume}{14}, \bibinfo{number}{7} (\bibinfo{year}{2010}),
  \bibinfo{pages}{645--662}.
\newblock


\bibitem[\protect\citeauthoryear{Fisher, Hammerla, Ploetz, Andras, Rochester,
  and Walker}{Fisher et~al\mbox{.}}{2016}]%
        {fisher2016unsupervised}
\bibfield{author}{\bibinfo{person}{James~M Fisher}, \bibinfo{person}{Nils~Y
  Hammerla}, \bibinfo{person}{Thomas Ploetz}, \bibinfo{person}{Peter Andras},
  \bibinfo{person}{Lynn Rochester}, {and} \bibinfo{person}{Richard~W Walker}.}
  \bibinfo{year}{2016}\natexlab{}.
\newblock \showarticletitle{Unsupervised home monitoring of Parkinson's disease
  motor symptoms using body-worn accelerometers}.
\newblock \bibinfo{journal}{\emph{Parkinsonism \& related disorders}}
  \bibinfo{volume}{33} (\bibinfo{year}{2016}), \bibinfo{pages}{44--50}.
\newblock


\bibitem[\protect\citeauthoryear{Gidaris, Singh, and Komodakis}{Gidaris
  et~al\mbox{.}}{2018}]%
        {gidaris2018unsupervised}
\bibfield{author}{\bibinfo{person}{Spyros Gidaris}, \bibinfo{person}{Praveer
  Singh}, {and} \bibinfo{person}{Nikos Komodakis}.}
  \bibinfo{year}{2018}\natexlab{}.
\newblock \showarticletitle{Unsupervised representation learning by predicting
  image rotations}.
\newblock \bibinfo{journal}{\emph{arXiv preprint arXiv:1803.07728}}
  (\bibinfo{year}{2018}).
\newblock


\bibitem[\protect\citeauthoryear{Graves, Mohamed, and Hinton}{Graves
  et~al\mbox{.}}{2013}]%
        {graves2013speech}
\bibfield{author}{\bibinfo{person}{Alex Graves}, \bibinfo{person}{Abdel-rahman
  Mohamed}, {and} \bibinfo{person}{Geoffrey Hinton}.}
  \bibinfo{year}{2013}\natexlab{}.
\newblock \showarticletitle{Speech recognition with deep recurrent neural
  networks}. In \bibinfo{booktitle}{\emph{2013 IEEE international conference on
  acoustics, speech and signal processing}}. IEEE, \bibinfo{pages}{6645--6649}.
\newblock


\bibitem[\protect\citeauthoryear{Guan and Pl{\"o}tz}{Guan and
  Pl{\"o}tz}{2017}]%
        {guan2017ensembles}
\bibfield{author}{\bibinfo{person}{Yu Guan} {and} \bibinfo{person}{Thomas
  Pl{\"o}tz}.} \bibinfo{year}{2017}\natexlab{}.
\newblock \showarticletitle{Ensembles of deep lstm learners for activity
  recognition using wearables}.
\newblock \bibinfo{journal}{\emph{Proceedings of the ACM on Interactive,
  Mobile, Wearable and Ubiquitous Technologies}} \bibinfo{volume}{1},
  \bibinfo{number}{2} (\bibinfo{year}{2017}), \bibinfo{pages}{1--28}.
\newblock


\bibitem[\protect\citeauthoryear{Gutmann and Hyv{\"a}rinen}{Gutmann and
  Hyv{\"a}rinen}{2010}]%
        {gutmann2010noise}
\bibfield{author}{\bibinfo{person}{Michael Gutmann} {and} \bibinfo{person}{Aapo
  Hyv{\"a}rinen}.} \bibinfo{year}{2010}\natexlab{}.
\newblock \showarticletitle{Noise-contrastive estimation: A new estimation
  principle for unnormalized statistical models}. In
  \bibinfo{booktitle}{\emph{Proceedings of the Thirteenth International
  Conference on Artificial Intelligence and Statistics}}.
  \bibinfo{pages}{297--304}.
\newblock


\bibitem[\protect\citeauthoryear{Hammerla, Halloran, and Pl{\"o}tz}{Hammerla
  et~al\mbox{.}}{2016}]%
        {hammerla2016deep}
\bibfield{author}{\bibinfo{person}{Nils~Y Hammerla}, \bibinfo{person}{Shane
  Halloran}, {and} \bibinfo{person}{Thomas Pl{\"o}tz}.}
  \bibinfo{year}{2016}\natexlab{}.
\newblock \showarticletitle{Deep, convolutional, and recurrent models for human
  activity recognition using wearables}.
\newblock \bibinfo{journal}{\emph{arXiv preprint arXiv:1604.08880}}
  (\bibinfo{year}{2016}).
\newblock


\bibitem[\protect\citeauthoryear{Hammerla, Kirkham, Andras, and
  Ploetz}{Hammerla et~al\mbox{.}}{2013}]%
        {hammerla2013preserving}
\bibfield{author}{\bibinfo{person}{Nils~Y Hammerla}, \bibinfo{person}{Reuben
  Kirkham}, \bibinfo{person}{Peter Andras}, {and} \bibinfo{person}{Thomas
  Ploetz}.} \bibinfo{year}{2013}\natexlab{}.
\newblock \showarticletitle{On preserving statistical characteristics of
  accelerometry data using their empirical cumulative distribution}. In
  \bibinfo{booktitle}{\emph{Proceedings of the 2013 international symposium on
  wearable computers}}. \bibinfo{pages}{65--68}.
\newblock


\bibitem[\protect\citeauthoryear{Haresamudram, Anderson, and
  Pl{\"o}tz}{Haresamudram et~al\mbox{.}}{2019}]%
        {haresamudram2019role}
\bibfield{author}{\bibinfo{person}{Harish Haresamudram},
  \bibinfo{person}{David~V Anderson}, {and} \bibinfo{person}{Thomas
  Pl{\"o}tz}.} \bibinfo{year}{2019}\natexlab{}.
\newblock \showarticletitle{On the role of features in human activity
  recognition}. In \bibinfo{booktitle}{\emph{Proceedings of the 23rd
  International Symposium on Wearable Computers}}. \bibinfo{pages}{78--88}.
\newblock


\bibitem[\protect\citeauthoryear{Haresamudram, Beedu, Agrawal, Grady, Essa,
  Hoffman, and Pl{\"o}tz}{Haresamudram et~al\mbox{.}}{2020}]%
        {haresamudram2020masked}
\bibfield{author}{\bibinfo{person}{Harish Haresamudram},
  \bibinfo{person}{Apoorva Beedu}, \bibinfo{person}{Varun Agrawal},
  \bibinfo{person}{Patrick~L Grady}, \bibinfo{person}{Irfan Essa},
  \bibinfo{person}{Judy Hoffman}, {and} \bibinfo{person}{Thomas Pl{\"o}tz}.}
  \bibinfo{year}{2020}\natexlab{}.
\newblock \showarticletitle{Masked reconstruction based self-supervision for
  human activity recognition}. In \bibinfo{booktitle}{\emph{Proceedings of the
  2020 International Symposium on Wearable Computers}}.
  \bibinfo{pages}{45--49}.
\newblock


\bibitem[\protect\citeauthoryear{He, Fan, Wu, Xie, and Girshick}{He
  et~al\mbox{.}}{2020}]%
        {he2020momentum}
\bibfield{author}{\bibinfo{person}{Kaiming He}, \bibinfo{person}{Haoqi Fan},
  \bibinfo{person}{Yuxin Wu}, \bibinfo{person}{Saining Xie}, {and}
  \bibinfo{person}{Ross Girshick}.} \bibinfo{year}{2020}\natexlab{}.
\newblock \showarticletitle{Momentum contrast for unsupervised visual
  representation learning}. In \bibinfo{booktitle}{\emph{Proceedings of the
  IEEE/CVF Conference on Computer Vision and Pattern Recognition}}.
  \bibinfo{pages}{9729--9738}.
\newblock


\bibitem[\protect\citeauthoryear{He, Zhang, Ren, and Sun}{He
  et~al\mbox{.}}{2016}]%
        {he2016deep}
\bibfield{author}{\bibinfo{person}{Kaiming He}, \bibinfo{person}{Xiangyu
  Zhang}, \bibinfo{person}{Shaoqing Ren}, {and} \bibinfo{person}{Jian Sun}.}
  \bibinfo{year}{2016}\natexlab{}.
\newblock \showarticletitle{Deep residual learning for image recognition}. In
  \bibinfo{booktitle}{\emph{Proceedings of the IEEE conference on computer
  vision and pattern recognition}}. \bibinfo{pages}{770--778}.
\newblock


\bibitem[\protect\citeauthoryear{H{\'e}naff, Srinivas, De~Fauw, Razavi,
  Doersch, Eslami, and Oord}{H{\'e}naff et~al\mbox{.}}{2019}]%
        {henaff2019data}
\bibfield{author}{\bibinfo{person}{Olivier~J H{\'e}naff},
  \bibinfo{person}{Aravind Srinivas}, \bibinfo{person}{Jeffrey De~Fauw},
  \bibinfo{person}{Ali Razavi}, \bibinfo{person}{Carl Doersch},
  \bibinfo{person}{SM Eslami}, {and} \bibinfo{person}{Aaron van~den Oord}.}
  \bibinfo{year}{2019}\natexlab{}.
\newblock \showarticletitle{Data-efficient image recognition with contrastive
  predictive coding}.
\newblock \bibinfo{journal}{\emph{arXiv preprint arXiv:1905.09272}}
  (\bibinfo{year}{2019}).
\newblock


\bibitem[\protect\citeauthoryear{Hochreiter and Schmidhuber}{Hochreiter and
  Schmidhuber}{1997}]%
        {hochreiter1997long}
\bibfield{author}{\bibinfo{person}{Sepp Hochreiter} {and}
  \bibinfo{person}{J{\"u}rgen Schmidhuber}.} \bibinfo{year}{1997}\natexlab{}.
\newblock \showarticletitle{Long short-term memory}.
\newblock \bibinfo{journal}{\emph{Neural computation}} \bibinfo{volume}{9},
  \bibinfo{number}{8} (\bibinfo{year}{1997}), \bibinfo{pages}{1735--1780}.
\newblock


\bibitem[\protect\citeauthoryear{Huang, Kaufmann, Aksan, Black, Hilliges, and
  Pons-Moll}{Huang et~al\mbox{.}}{2018}]%
        {huang2018deep}
\bibfield{author}{\bibinfo{person}{Yinghao Huang}, \bibinfo{person}{Manuel
  Kaufmann}, \bibinfo{person}{Emre Aksan}, \bibinfo{person}{Michael~J Black},
  \bibinfo{person}{Otmar Hilliges}, {and} \bibinfo{person}{Gerard Pons-Moll}.}
  \bibinfo{year}{2018}\natexlab{}.
\newblock \showarticletitle{Deep inertial poser: learning to reconstruct human
  pose from sparse inertial measurements in real time}.
\newblock \bibinfo{journal}{\emph{ACM Transactions on Graphics (TOG)}}
  \bibinfo{volume}{37}, \bibinfo{number}{6} (\bibinfo{year}{2018}),
  \bibinfo{pages}{1--15}.
\newblock


\bibitem[\protect\citeauthoryear{Huynh and Schiele}{Huynh and Schiele}{2005}]%
        {huynh2005analyzing}
\bibfield{author}{\bibinfo{person}{T{\^a}m Huynh} {and} \bibinfo{person}{Bernt
  Schiele}.} \bibinfo{year}{2005}\natexlab{}.
\newblock \showarticletitle{Analyzing features for activity recognition}. In
  \bibinfo{booktitle}{\emph{Proceedings of the 2005 joint conference on Smart
  objects and ambient intelligence: innovative context-aware services: usages
  and technologies}}. \bibinfo{pages}{159--163}.
\newblock


\bibitem[\protect\citeauthoryear{Hyvarinen and Morioka}{Hyvarinen and
  Morioka}{2016}]%
        {hyvarinen2016unsupervised}
\bibfield{author}{\bibinfo{person}{Aapo Hyvarinen} {and}
  \bibinfo{person}{Hiroshi Morioka}.} \bibinfo{year}{2016}\natexlab{}.
\newblock \showarticletitle{Unsupervised feature extraction by time-contrastive
  learning and nonlinear ICA}. In \bibinfo{booktitle}{\emph{Advances in Neural
  Information Processing Systems}}. \bibinfo{pages}{3765--3773}.
\newblock


\bibitem[\protect\citeauthoryear{Ioffe and Szegedy}{Ioffe and Szegedy}{2015}]%
        {ioffe2015batch}
\bibfield{author}{\bibinfo{person}{Sergey Ioffe} {and}
  \bibinfo{person}{Christian Szegedy}.} \bibinfo{year}{2015}\natexlab{}.
\newblock \showarticletitle{Batch normalization: Accelerating deep network
  training by reducing internal covariate shift}.
\newblock \bibinfo{journal}{\emph{arXiv preprint arXiv:1502.03167}}
  (\bibinfo{year}{2015}).
\newblock


\bibitem[\protect\citeauthoryear{Kingma and Ba}{Kingma and Ba}{2014}]%
        {kingma2014adam}
\bibfield{author}{\bibinfo{person}{Diederik~P Kingma} {and}
  \bibinfo{person}{Jimmy Ba}.} \bibinfo{year}{2014}\natexlab{}.
\newblock \showarticletitle{Adam: A method for stochastic optimization}.
\newblock \bibinfo{journal}{\emph{arXiv preprint arXiv:1412.6980}}
  (\bibinfo{year}{2014}).
\newblock


\bibitem[\protect\citeauthoryear{Krizhevsky, Sutskever, and Hinton}{Krizhevsky
  et~al\mbox{.}}{2017}]%
        {krizhevsky2017imagenet}
\bibfield{author}{\bibinfo{person}{Alex Krizhevsky}, \bibinfo{person}{Ilya
  Sutskever}, {and} \bibinfo{person}{Geoffrey~E Hinton}.}
  \bibinfo{year}{2017}\natexlab{}.
\newblock \showarticletitle{Imagenet classification with deep convolutional
  neural networks}.
\newblock \bibinfo{journal}{\emph{Commun. ACM}} \bibinfo{volume}{60},
  \bibinfo{number}{6} (\bibinfo{year}{2017}), \bibinfo{pages}{84--90}.
\newblock


\bibitem[\protect\citeauthoryear{Kwon, Abowd, and Pl{\"o}tz}{Kwon
  et~al\mbox{.}}{2018}]%
        {kwon2018adding}
\bibfield{author}{\bibinfo{person}{Hyeokhyen Kwon}, \bibinfo{person}{Gregory~D
  Abowd}, {and} \bibinfo{person}{Thomas Pl{\"o}tz}.}
  \bibinfo{year}{2018}\natexlab{}.
\newblock \showarticletitle{Adding structural characteristics to
  distribution-based accelerometer representations for activity recognition
  using wearables}. In \bibinfo{booktitle}{\emph{Proceedings of the 2018 ACM
  international symposium on wearable computers}}. \bibinfo{pages}{72--75}.
\newblock


\bibitem[\protect\citeauthoryear{Kwon, Tong, Haresamudram, Gao, Abowd, Lane,
  and Ploetz}{Kwon et~al\mbox{.}}{2020}]%
        {kwon2020imutube}
\bibfield{author}{\bibinfo{person}{Hyeokhyen Kwon}, \bibinfo{person}{Catherine
  Tong}, \bibinfo{person}{Harish Haresamudram}, \bibinfo{person}{Yan Gao},
  \bibinfo{person}{Gregory~D Abowd}, \bibinfo{person}{Nicholas~D Lane}, {and}
  \bibinfo{person}{Thomas Ploetz}.} \bibinfo{year}{2020}\natexlab{}.
\newblock \showarticletitle{IMUTube: Automatic extraction of virtual on-body
  accelerometry from video for human activity recognition}.
\newblock \bibinfo{journal}{\emph{arXiv preprint arXiv:2006.05675}}
  (\bibinfo{year}{2020}).
\newblock


\bibitem[\protect\citeauthoryear{LeCun, Bengio, and Hinton}{LeCun
  et~al\mbox{.}}{2015}]%
        {lecun2015deep}
\bibfield{author}{\bibinfo{person}{Yann LeCun}, \bibinfo{person}{Yoshua
  Bengio}, {and} \bibinfo{person}{Geoffrey Hinton}.}
  \bibinfo{year}{2015}\natexlab{}.
\newblock \showarticletitle{Deep learning}.
\newblock \bibinfo{journal}{\emph{nature}} \bibinfo{volume}{521},
  \bibinfo{number}{7553} (\bibinfo{year}{2015}), \bibinfo{pages}{436--444}.
\newblock


\bibitem[\protect\citeauthoryear{Li, Abowd, and Pl{\"o}tz}{Li
  et~al\mbox{.}}{2018}]%
        {li2018specialized}
\bibfield{author}{\bibinfo{person}{Hong Li}, \bibinfo{person}{Gregory~D Abowd},
  {and} \bibinfo{person}{Thomas Pl{\"o}tz}.} \bibinfo{year}{2018}\natexlab{}.
\newblock \showarticletitle{On specialized window lengths and detector based
  human activity recognition}. In \bibinfo{booktitle}{\emph{Proceedings of the
  2018 ACM international symposium on wearable computers}}.
  \bibinfo{pages}{68--71}.
\newblock


\bibitem[\protect\citeauthoryear{Liu, Li, and Lee}{Liu et~al\mbox{.}}{2020a}]%
        {liu2020tera}
\bibfield{author}{\bibinfo{person}{Andy~T Liu}, \bibinfo{person}{Shang-Wen Li},
  {and} \bibinfo{person}{Hung-yi Lee}.} \bibinfo{year}{2020}\natexlab{a}.
\newblock \showarticletitle{Tera: Self-supervised learning of transformer
  encoder representation for speech}.
\newblock \bibinfo{journal}{\emph{arXiv preprint arXiv:2007.06028}}
  (\bibinfo{year}{2020}).
\newblock


\bibitem[\protect\citeauthoryear{Liu, Yang, Chi, Hsu, and Lee}{Liu
  et~al\mbox{.}}{2020b}]%
        {liu2020mockingjay}
\bibfield{author}{\bibinfo{person}{Andy~T Liu}, \bibinfo{person}{Shu-wen Yang},
  \bibinfo{person}{Po-Han Chi}, \bibinfo{person}{Po-chun Hsu}, {and}
  \bibinfo{person}{Hung-yi Lee}.} \bibinfo{year}{2020}\natexlab{b}.
\newblock \showarticletitle{Mockingjay: Unsupervised speech representation
  learning with deep bidirectional transformer encoders}. In
  \bibinfo{booktitle}{\emph{ICASSP 2020-2020 IEEE International Conference on
  Acoustics, Speech and Signal Processing (ICASSP)}}. IEEE,
  \bibinfo{pages}{6419--6423}.
\newblock


\bibitem[\protect\citeauthoryear{Logeswaran and Lee}{Logeswaran and
  Lee}{2018}]%
        {logeswaran2018efficient}
\bibfield{author}{\bibinfo{person}{Lajanugen Logeswaran} {and}
  \bibinfo{person}{Honglak Lee}.} \bibinfo{year}{2018}\natexlab{}.
\newblock \showarticletitle{An efficient framework for learning sentence
  representations}.
\newblock \bibinfo{journal}{\emph{arXiv preprint arXiv:1803.02893}}
  (\bibinfo{year}{2018}).
\newblock


\bibitem[\protect\citeauthoryear{Mahmud, Tonmoy, Bhaumik, Rahman, Amin,
  Shoyaib, Khan, and Ali}{Mahmud et~al\mbox{.}}{2020}]%
        {mahmud2020human}
\bibfield{author}{\bibinfo{person}{Saif Mahmud}, \bibinfo{person}{M Tonmoy},
  \bibinfo{person}{Kishor~Kumar Bhaumik}, \bibinfo{person}{AKM Rahman},
  \bibinfo{person}{M~Ashraful Amin}, \bibinfo{person}{Mohammad Shoyaib},
  \bibinfo{person}{Muhammad Asif~Hossain Khan}, {and}
  \bibinfo{person}{Amin~Ahsan Ali}.} \bibinfo{year}{2020}\natexlab{}.
\newblock \showarticletitle{Human Activity Recognition from Wearable Sensor
  Data Using Self-Attention}.
\newblock \bibinfo{journal}{\emph{arXiv preprint arXiv:2003.09018}}
  (\bibinfo{year}{2020}).
\newblock


\bibitem[\protect\citeauthoryear{Malekzadeh, Clegg, Cavallaro, and
  Haddadi}{Malekzadeh et~al\mbox{.}}{2018}]%
        {malekzadeh2018protecting}
\bibfield{author}{\bibinfo{person}{Mohammad Malekzadeh},
  \bibinfo{person}{Richard~G Clegg}, \bibinfo{person}{Andrea Cavallaro}, {and}
  \bibinfo{person}{Hamed Haddadi}.} \bibinfo{year}{2018}\natexlab{}.
\newblock \showarticletitle{Protecting sensory data against sensitive
  inferences}. In \bibinfo{booktitle}{\emph{Proceedings of the 1st Workshop on
  Privacy by Design in Distributed Systems}}. \bibinfo{pages}{1--6}.
\newblock


\bibitem[\protect\citeauthoryear{Mikolov, Sutskever, Chen, Corrado, and
  Dean}{Mikolov et~al\mbox{.}}{2013}]%
        {mikolov2013distributed}
\bibfield{author}{\bibinfo{person}{Tomas Mikolov}, \bibinfo{person}{Ilya
  Sutskever}, \bibinfo{person}{Kai Chen}, \bibinfo{person}{Greg~S Corrado},
  {and} \bibinfo{person}{Jeff Dean}.} \bibinfo{year}{2013}\natexlab{}.
\newblock \showarticletitle{Distributed representations of words and phrases
  and their compositionality}. In \bibinfo{booktitle}{\emph{Advances in neural
  information processing systems}}. \bibinfo{pages}{3111--3119}.
\newblock


\bibitem[\protect\citeauthoryear{Misra, Zitnick, and Hebert}{Misra
  et~al\mbox{.}}{2016}]%
        {misra2016shuffle}
\bibfield{author}{\bibinfo{person}{Ishan Misra}, \bibinfo{person}{C~Lawrence
  Zitnick}, {and} \bibinfo{person}{Martial Hebert}.}
  \bibinfo{year}{2016}\natexlab{}.
\newblock \showarticletitle{Shuffle and learn: unsupervised learning using
  temporal order verification}. In \bibinfo{booktitle}{\emph{European
  Conference on Computer Vision}}. Springer, \bibinfo{pages}{527--544}.
\newblock


\bibitem[\protect\citeauthoryear{Morshed, Saha, Li, D'Mello, De~Choudhury,
  Abowd, and Pl{\"o}tz}{Morshed et~al\mbox{.}}{2019}]%
        {morshed2019prediction}
\bibfield{author}{\bibinfo{person}{Mehrab~Bin Morshed},
  \bibinfo{person}{Koustuv Saha}, \bibinfo{person}{Richard Li},
  \bibinfo{person}{Sidney~K D'Mello}, \bibinfo{person}{Munmun De~Choudhury},
  \bibinfo{person}{Gregory~D Abowd}, {and} \bibinfo{person}{Thomas Pl{\"o}tz}.}
  \bibinfo{year}{2019}\natexlab{}.
\newblock \showarticletitle{Prediction of mood instability with passive
  sensing}.
\newblock \bibinfo{journal}{\emph{Proceedings of the ACM on Interactive,
  Mobile, Wearable and Ubiquitous Technologies}} \bibinfo{volume}{3},
  \bibinfo{number}{3} (\bibinfo{year}{2019}), \bibinfo{pages}{1--21}.
\newblock


\bibitem[\protect\citeauthoryear{Murahari and Pl{\"o}tz}{Murahari and
  Pl{\"o}tz}{2018}]%
        {murahari2018attention}
\bibfield{author}{\bibinfo{person}{Vishvak~S Murahari} {and}
  \bibinfo{person}{Thomas Pl{\"o}tz}.} \bibinfo{year}{2018}\natexlab{}.
\newblock \showarticletitle{On attention models for human activity
  recognition}. In \bibinfo{booktitle}{\emph{Proceedings of the 2018 ACM
  International Symposium on Wearable Computers}}. \bibinfo{pages}{100--103}.
\newblock


\bibitem[\protect\citeauthoryear{Nair and Hinton}{Nair and Hinton}{2010}]%
        {nair2010rectified}
\bibfield{author}{\bibinfo{person}{Vinod Nair} {and}
  \bibinfo{person}{Geoffrey~E Hinton}.} \bibinfo{year}{2010}\natexlab{}.
\newblock \showarticletitle{Rectified linear units improve restricted boltzmann
  machines}. In \bibinfo{booktitle}{\emph{ICML}}.
\newblock


\bibitem[\protect\citeauthoryear{Nutt, Bloem, Giladi, Hallett, Horak, and
  Nieuwboer}{Nutt et~al\mbox{.}}{2011}]%
        {nutt2011freezing}
\bibfield{author}{\bibinfo{person}{John~G Nutt}, \bibinfo{person}{Bastiaan~R
  Bloem}, \bibinfo{person}{Nir Giladi}, \bibinfo{person}{Mark Hallett},
  \bibinfo{person}{Fay~B Horak}, {and} \bibinfo{person}{Alice Nieuwboer}.}
  \bibinfo{year}{2011}\natexlab{}.
\newblock \showarticletitle{Freezing of gait: moving forward on a mysterious
  clinical phenomenon}.
\newblock \bibinfo{journal}{\emph{The Lancet Neurology}} \bibinfo{volume}{10},
  \bibinfo{number}{8} (\bibinfo{year}{2011}), \bibinfo{pages}{734--744}.
\newblock


\bibitem[\protect\citeauthoryear{Oord, Li, and Vinyals}{Oord
  et~al\mbox{.}}{2018}]%
        {oord2018representation}
\bibfield{author}{\bibinfo{person}{Aaron van~den Oord}, \bibinfo{person}{Yazhe
  Li}, {and} \bibinfo{person}{Oriol Vinyals}.} \bibinfo{year}{2018}\natexlab{}.
\newblock \showarticletitle{Representation learning with contrastive predictive
  coding}.
\newblock \bibinfo{journal}{\emph{arXiv preprint arXiv:1807.03748}}
  (\bibinfo{year}{2018}).
\newblock


\bibitem[\protect\citeauthoryear{Ord{\'o}{\~n}ez and Roggen}{Ord{\'o}{\~n}ez
  and Roggen}{2016}]%
        {ordonez2016deep}
\bibfield{author}{\bibinfo{person}{Francisco~Javier Ord{\'o}{\~n}ez} {and}
  \bibinfo{person}{Daniel Roggen}.} \bibinfo{year}{2016}\natexlab{}.
\newblock \showarticletitle{Deep convolutional and lstm recurrent neural
  networks for multimodal wearable activity recognition}.
\newblock \bibinfo{journal}{\emph{Sensors}} \bibinfo{volume}{16},
  \bibinfo{number}{1} (\bibinfo{year}{2016}), \bibinfo{pages}{115}.
\newblock


\bibitem[\protect\citeauthoryear{Paszke, Gross, Massa, Lerer, Bradbury, Chanan,
  Killeen, Lin, Gimelshein, Antiga, et~al\mbox{.}}{Paszke
  et~al\mbox{.}}{2019}]%
        {paszke2019pytorch}
\bibfield{author}{\bibinfo{person}{Adam Paszke}, \bibinfo{person}{Sam Gross},
  \bibinfo{person}{Francisco Massa}, \bibinfo{person}{Adam Lerer},
  \bibinfo{person}{James Bradbury}, \bibinfo{person}{Gregory Chanan},
  \bibinfo{person}{Trevor Killeen}, \bibinfo{person}{Zeming Lin},
  \bibinfo{person}{Natalia Gimelshein}, \bibinfo{person}{Luca Antiga},
  {et~al\mbox{.}}} \bibinfo{year}{2019}\natexlab{}.
\newblock \showarticletitle{Pytorch: An imperative style, high-performance deep
  learning library}. In \bibinfo{booktitle}{\emph{Advances in neural
  information processing systems}}. \bibinfo{pages}{8026--8037}.
\newblock


\bibitem[\protect\citeauthoryear{Pathak, Krahenbuhl, Donahue, Darrell, and
  Efros}{Pathak et~al\mbox{.}}{2016}]%
        {pathak2016context}
\bibfield{author}{\bibinfo{person}{Deepak Pathak}, \bibinfo{person}{Philipp
  Krahenbuhl}, \bibinfo{person}{Jeff Donahue}, \bibinfo{person}{Trevor
  Darrell}, {and} \bibinfo{person}{Alexei~A Efros}.}
  \bibinfo{year}{2016}\natexlab{}.
\newblock \showarticletitle{Context encoders: Feature learning by inpainting}.
  In \bibinfo{booktitle}{\emph{Proceedings of the IEEE conference on computer
  vision and pattern recognition}}. \bibinfo{pages}{2536--2544}.
\newblock


\bibitem[\protect\citeauthoryear{Pennington, Socher, and Manning}{Pennington
  et~al\mbox{.}}{2014}]%
        {pennington2014glove}
\bibfield{author}{\bibinfo{person}{Jeffrey Pennington},
  \bibinfo{person}{Richard Socher}, {and} \bibinfo{person}{Christopher~D
  Manning}.} \bibinfo{year}{2014}\natexlab{}.
\newblock \showarticletitle{Glove: Global vectors for word representation}. In
  \bibinfo{booktitle}{\emph{Proceedings of the 2014 conference on empirical
  methods in natural language processing (EMNLP)}}.
  \bibinfo{pages}{1532--1543}.
\newblock


\bibitem[\protect\citeauthoryear{Pl{\"o}tz, Hammerla, and Olivier}{Pl{\"o}tz
  et~al\mbox{.}}{2011}]%
        {plotz2011feature}
\bibfield{author}{\bibinfo{person}{Thomas Pl{\"o}tz}, \bibinfo{person}{Nils~Y
  Hammerla}, {and} \bibinfo{person}{Patrick~L Olivier}.}
  \bibinfo{year}{2011}\natexlab{}.
\newblock \showarticletitle{Feature learning for activity recognition in
  ubiquitous computing}. In \bibinfo{booktitle}{\emph{Twenty-second
  international joint conference on artificial intelligence}}.
\newblock


\bibitem[\protect\citeauthoryear{Powers}{Powers}{2020}]%
        {powers2020evaluation}
\bibfield{author}{\bibinfo{person}{David~MW Powers}.}
  \bibinfo{year}{2020}\natexlab{}.
\newblock \showarticletitle{Evaluation: from precision, recall and F-measure to
  ROC, informedness, markedness and correlation}.
\newblock \bibinfo{journal}{\emph{arXiv preprint arXiv:2010.16061}}
  (\bibinfo{year}{2020}).
\newblock


\bibitem[\protect\citeauthoryear{Radford, Narasimhan, Salimans, and
  Sutskever}{Radford et~al\mbox{.}}{2018}]%
        {radford2018improving}
\bibfield{author}{\bibinfo{person}{Alec Radford}, \bibinfo{person}{Karthik
  Narasimhan}, \bibinfo{person}{Tim Salimans}, {and} \bibinfo{person}{Ilya
  Sutskever}.} \bibinfo{year}{2018}\natexlab{}.
\newblock \bibinfo{title}{Improving language understanding by generative
  pre-training}.
\newblock
\newblock


\bibitem[\protect\citeauthoryear{Reiss and Stricker}{Reiss and
  Stricker}{2012}]%
        {reiss2012introducing}
\bibfield{author}{\bibinfo{person}{A. Reiss} {and} \bibinfo{person}{D.
  Stricker}.} \bibinfo{year}{2012}\natexlab{}.
\newblock \showarticletitle{Introducing a new benchmarked dataset for activity
  monitoring}.
\newblock


\bibitem[\protect\citeauthoryear{Reyes, Zhang, Ghosh, Shah, Wu, Parnami,
  Bercik, Starner, Abowd, and Edwards}{Reyes et~al\mbox{.}}{2016}]%
        {reyes2016whoosh}
\bibfield{author}{\bibinfo{person}{Gabriel Reyes}, \bibinfo{person}{Dingtian
  Zhang}, \bibinfo{person}{Sarthak Ghosh}, \bibinfo{person}{Pratik Shah},
  \bibinfo{person}{Jason Wu}, \bibinfo{person}{Aman Parnami},
  \bibinfo{person}{Bailey Bercik}, \bibinfo{person}{Thad Starner},
  \bibinfo{person}{Gregory~D Abowd}, {and} \bibinfo{person}{W~Keith Edwards}.}
  \bibinfo{year}{2016}\natexlab{}.
\newblock \showarticletitle{Whoosh: non-voice acoustics for low-cost,
  hands-free, and rapid input on smartwatches}. In
  \bibinfo{booktitle}{\emph{Proceedings of the 2016 ACM International Symposium
  on Wearable Computers}}. \bibinfo{pages}{120--127}.
\newblock


\bibitem[\protect\citeauthoryear{Rivi{\`e}re, Joulin, Mazar{\'e}, and
  Dupoux}{Rivi{\`e}re et~al\mbox{.}}{2020}]%
        {riviere2020unsupervised}
\bibfield{author}{\bibinfo{person}{Morgane Rivi{\`e}re},
  \bibinfo{person}{Armand Joulin}, \bibinfo{person}{Pierre-Emmanuel
  Mazar{\'e}}, {and} \bibinfo{person}{Emmanuel Dupoux}.}
  \bibinfo{year}{2020}\natexlab{}.
\newblock \showarticletitle{Unsupervised pretraining transfers well across
  languages}. In \bibinfo{booktitle}{\emph{ICASSP 2020-2020 IEEE International
  Conference on Acoustics, Speech and Signal Processing (ICASSP)}}. IEEE,
  \bibinfo{pages}{7414--7418}.
\newblock


\bibitem[\protect\citeauthoryear{Saeed, Ozcelebi, and Lukkien}{Saeed
  et~al\mbox{.}}{2019}]%
        {saeed2019multi}
\bibfield{author}{\bibinfo{person}{Aaqib Saeed}, \bibinfo{person}{Tanir
  Ozcelebi}, {and} \bibinfo{person}{Johan Lukkien}.}
  \bibinfo{year}{2019}\natexlab{}.
\newblock \showarticletitle{Multi-task Self-Supervised Learning for Human
  Activity Detection}.
\newblock \bibinfo{journal}{\emph{Proceedings of the ACM on Interactive,
  Mobile, Wearable and Ubiquitous Technologies}} \bibinfo{volume}{3},
  \bibinfo{number}{2} (\bibinfo{year}{2019}), \bibinfo{pages}{1--30}.
\newblock


\bibitem[\protect\citeauthoryear{Srivastava, Hinton, Krizhevsky, Sutskever, and
  Salakhutdinov}{Srivastava et~al\mbox{.}}{2014}]%
        {srivastava2014dropout}
\bibfield{author}{\bibinfo{person}{Nitish Srivastava},
  \bibinfo{person}{Geoffrey Hinton}, \bibinfo{person}{Alex Krizhevsky},
  \bibinfo{person}{Ilya Sutskever}, {and} \bibinfo{person}{Ruslan
  Salakhutdinov}.} \bibinfo{year}{2014}\natexlab{}.
\newblock \showarticletitle{Dropout: a simple way to prevent neural networks
  from overfitting}.
\newblock \bibinfo{journal}{\emph{The journal of machine learning research}}
  \bibinfo{volume}{15}, \bibinfo{number}{1} (\bibinfo{year}{2014}),
  \bibinfo{pages}{1929--1958}.
\newblock


\bibitem[\protect\citeauthoryear{Stiefmeier, Roggen, Ogris, Lukowicz, and
  Tr{\"o}ster}{Stiefmeier et~al\mbox{.}}{2008}]%
        {stiefmeier2008wearable}
\bibfield{author}{\bibinfo{person}{T. Stiefmeier}, \bibinfo{person}{D. Roggen},
  \bibinfo{person}{G. Ogris}, \bibinfo{person}{P. Lukowicz}, {and}
  \bibinfo{person}{G. Tr{\"o}ster}.} \bibinfo{year}{2008}\natexlab{}.
\newblock \showarticletitle{Wearable activity tracking in car manufacturing}.
\newblock \bibinfo{journal}{\emph{IEEE Pervasive Computing}}
  \bibinfo{number}{2} (\bibinfo{year}{2008}), \bibinfo{pages}{42--50}.
\newblock


\bibitem[\protect\citeauthoryear{Takeda, Okita, Lago, and Inoue}{Takeda
  et~al\mbox{.}}{2018}]%
        {takeda2018multi}
\bibfield{author}{\bibinfo{person}{Shingo Takeda}, \bibinfo{person}{Tsuyoshi
  Okita}, \bibinfo{person}{Paula Lago}, {and} \bibinfo{person}{Sozo Inoue}.}
  \bibinfo{year}{2018}\natexlab{}.
\newblock \showarticletitle{A multi-sensor setting activity recognition
  simulation tool}. In \bibinfo{booktitle}{\emph{Proceedings of the 2018 ACM
  International Joint Conference and 2018 International Symposium on Pervasive
  and Ubiquitous Computing and Wearable Computers}}.
  \bibinfo{pages}{1444--1448}.
\newblock


\bibitem[\protect\citeauthoryear{Thomaz, Essa, and Abowd}{Thomaz
  et~al\mbox{.}}{2015}]%
        {edison_eating}
\bibfield{author}{\bibinfo{person}{Edison Thomaz}, \bibinfo{person}{Irfan
  Essa}, {and} \bibinfo{person}{Gregory~D. Abowd}.}
  \bibinfo{year}{2015}\natexlab{}.
\newblock \showarticletitle{A Practical Approach for Recognizing Eating Moments
  with Wrist-Mounted Inertial Sensing}. In
  \bibinfo{booktitle}{\emph{Proceedings of the 2015 ACM International Joint
  Conference on Pervasive and Ubiquitous Computing}} (Osaka, Japan)
  \emph{(\bibinfo{series}{UbiComp '15})}. \bibinfo{publisher}{Association for
  Computing Machinery}, \bibinfo{address}{New York, NY, USA},
  \bibinfo{pages}{1029–1040}.
\newblock
\showISBNx{9781450335744}
\urldef\tempurl%
\url{https://doi.org/10.1145/2750858.2807545}
\showDOI{\tempurl}


\bibitem[\protect\citeauthoryear{Varamin, Abbasnejad, Shi, Ranasinghe, and
  Rezatofighi}{Varamin et~al\mbox{.}}{2018}]%
        {varamin2018deep}
\bibfield{author}{\bibinfo{person}{Alireza~Abedin Varamin},
  \bibinfo{person}{Ehsan Abbasnejad}, \bibinfo{person}{Qinfeng Shi},
  \bibinfo{person}{Damith~C Ranasinghe}, {and} \bibinfo{person}{Hamid
  Rezatofighi}.} \bibinfo{year}{2018}\natexlab{}.
\newblock \showarticletitle{Deep auto-set: A deep auto-encoder-set network for
  activity recognition using wearables}. In
  \bibinfo{booktitle}{\emph{Proceedings of the 15th EAI International
  Conference on Mobile and Ubiquitous Systems: Computing, Networking and
  Services}}. \bibinfo{pages}{246--253}.
\newblock


\bibitem[\protect\citeauthoryear{Vaswani, Shazeer, Parmar, Uszkoreit, Jones,
  Gomez, Kaiser, and Polosukhin}{Vaswani et~al\mbox{.}}{2017}]%
        {vaswani2017attention}
\bibfield{author}{\bibinfo{person}{Ashish Vaswani}, \bibinfo{person}{Noam
  Shazeer}, \bibinfo{person}{Niki Parmar}, \bibinfo{person}{Jakob Uszkoreit},
  \bibinfo{person}{Llion Jones}, \bibinfo{person}{Aidan~N Gomez},
  \bibinfo{person}{{\L}ukasz Kaiser}, {and} \bibinfo{person}{Illia
  Polosukhin}.} \bibinfo{year}{2017}\natexlab{}.
\newblock \showarticletitle{Attention is all you need}. In
  \bibinfo{booktitle}{\emph{Advances in neural information processing
  systems}}. \bibinfo{pages}{5998--6008}.
\newblock


\bibitem[\protect\citeauthoryear{Wang, Chen, Chen, Li, Harari, Tignor, Zhou,
  Ben-Zeev, and Campbell}{Wang et~al\mbox{.}}{2014}]%
        {wang2014studentlife}
\bibfield{author}{\bibinfo{person}{Rui Wang}, \bibinfo{person}{Fanglin Chen},
  \bibinfo{person}{Zhenyu Chen}, \bibinfo{person}{Tianxing Li},
  \bibinfo{person}{Gabriella Harari}, \bibinfo{person}{Stefanie Tignor},
  \bibinfo{person}{Xia Zhou}, \bibinfo{person}{Dror Ben-Zeev}, {and}
  \bibinfo{person}{Andrew~T Campbell}.} \bibinfo{year}{2014}\natexlab{}.
\newblock \showarticletitle{StudentLife: assessing mental health, academic
  performance and behavioral trends of college students using smartphones}. In
  \bibinfo{booktitle}{\emph{Proceedings of the 2014 ACM international joint
  conference on pervasive and ubiquitous computing}}. \bibinfo{pages}{3--14}.
\newblock


\bibitem[\protect\citeauthoryear{Wang, Tang, and Livescu}{Wang
  et~al\mbox{.}}{2020}]%
        {wang2020unsupervised}
\bibfield{author}{\bibinfo{person}{Weiran Wang}, \bibinfo{person}{Qingming
  Tang}, {and} \bibinfo{person}{Karen Livescu}.}
  \bibinfo{year}{2020}\natexlab{}.
\newblock \showarticletitle{Unsupervised pre-training of bidirectional speech
  encoders via masked reconstruction}. In \bibinfo{booktitle}{\emph{ICASSP
  2020-2020 IEEE International Conference on Acoustics, Speech and Signal
  Processing (ICASSP)}}. IEEE, \bibinfo{pages}{6889--6893}.
\newblock


\bibitem[\protect\citeauthoryear{Wei, Lim, Zisserman, and Freeman}{Wei
  et~al\mbox{.}}{2018}]%
        {wei2018learning}
\bibfield{author}{\bibinfo{person}{Donglai Wei}, \bibinfo{person}{Joseph~J
  Lim}, \bibinfo{person}{Andrew Zisserman}, {and} \bibinfo{person}{William~T
  Freeman}.} \bibinfo{year}{2018}\natexlab{}.
\newblock \showarticletitle{Learning and using the arrow of time}. In
  \bibinfo{booktitle}{\emph{Proceedings of the IEEE Conference on Computer
  Vision and Pattern Recognition}}. \bibinfo{pages}{8052--8060}.
\newblock


\bibitem[\protect\citeauthoryear{Yang, Nguyen, San, Li, and Krishnaswamy}{Yang
  et~al\mbox{.}}{2015}]%
        {yang2015deep}
\bibfield{author}{\bibinfo{person}{Jianbo Yang}, \bibinfo{person}{Minh~Nhut
  Nguyen}, \bibinfo{person}{Phyo~Phyo San}, \bibinfo{person}{Xiaoli Li}, {and}
  \bibinfo{person}{Shonali Krishnaswamy}.} \bibinfo{year}{2015}\natexlab{}.
\newblock \showarticletitle{Deep convolutional neural networks on multichannel
  time series for human activity recognition.}. In
  \bibinfo{booktitle}{\emph{Ijcai}}, Vol.~\bibinfo{volume}{15}. Citeseer,
  \bibinfo{pages}{3995--4001}.
\newblock


\bibitem[\protect\citeauthoryear{Zeng, Gao, Yu, Mengshoel, Langseth, Lane, and
  Liu}{Zeng et~al\mbox{.}}{2018}]%
        {zeng2018understanding}
\bibfield{author}{\bibinfo{person}{Ming Zeng}, \bibinfo{person}{Haoxiang Gao},
  \bibinfo{person}{Tong Yu}, \bibinfo{person}{Ole~J Mengshoel},
  \bibinfo{person}{Helge Langseth}, \bibinfo{person}{Ian Lane}, {and}
  \bibinfo{person}{Xiaobing Liu}.} \bibinfo{year}{2018}\natexlab{}.
\newblock \showarticletitle{Understanding and improving recurrent networks for
  human activity recognition by continuous attention}. In
  \bibinfo{booktitle}{\emph{Proceedings of the 2018 ACM International Symposium
  on Wearable Computers}}. \bibinfo{pages}{56--63}.
\newblock


\bibitem[\protect\citeauthoryear{Zeng, Nguyen, Yu, Mengshoel, Zhu, Wu, and
  Zhang}{Zeng et~al\mbox{.}}{2014}]%
        {zeng2014convolutional}
\bibfield{author}{\bibinfo{person}{Ming Zeng}, \bibinfo{person}{Le~T Nguyen},
  \bibinfo{person}{Bo Yu}, \bibinfo{person}{Ole~J Mengshoel},
  \bibinfo{person}{Jiang Zhu}, \bibinfo{person}{Pang Wu}, {and}
  \bibinfo{person}{Joy Zhang}.} \bibinfo{year}{2014}\natexlab{}.
\newblock \showarticletitle{Convolutional neural networks for human activity
  recognition using mobile sensors}. In \bibinfo{booktitle}{\emph{6th
  International Conference on Mobile Computing, Applications and Services}}.
  IEEE, \bibinfo{pages}{197--205}.
\newblock


\bibitem[\protect\citeauthoryear{Zhang, Bedri, Reyes, Bercik, Inan, Starner,
  and Abowd}{Zhang et~al\mbox{.}}{2016a}]%
        {zhang2016tapskin}
\bibfield{author}{\bibinfo{person}{Cheng Zhang}, \bibinfo{person}{AbdelKareem
  Bedri}, \bibinfo{person}{Gabriel Reyes}, \bibinfo{person}{Bailey Bercik},
  \bibinfo{person}{Omer~T Inan}, \bibinfo{person}{Thad~E Starner}, {and}
  \bibinfo{person}{Gregory~D Abowd}.} \bibinfo{year}{2016}\natexlab{a}.
\newblock \showarticletitle{Tapskin: Recognizing on-skin input for
  smartwatches}. In \bibinfo{booktitle}{\emph{Proceedings of the 2016 ACM
  International Conference on Interactive Surfaces and Spaces}}.
  \bibinfo{pages}{13--22}.
\newblock


\bibitem[\protect\citeauthoryear{Zhang, Waghmare, Kundra, Pu, Gilliland,
  Ploetz, Starner, Inan, and Abowd}{Zhang et~al\mbox{.}}{2017a}]%
        {zhang2017fingersound}
\bibfield{author}{\bibinfo{person}{Cheng Zhang}, \bibinfo{person}{Anandghan
  Waghmare}, \bibinfo{person}{Pranav Kundra}, \bibinfo{person}{Yiming Pu},
  \bibinfo{person}{Scott Gilliland}, \bibinfo{person}{Thomas Ploetz},
  \bibinfo{person}{Thad~E Starner}, \bibinfo{person}{Omer~T Inan}, {and}
  \bibinfo{person}{Gregory~D Abowd}.} \bibinfo{year}{2017}\natexlab{a}.
\newblock \showarticletitle{Fingersound: Recognizing unistroke thumb gestures
  using a ring}.
\newblock \bibinfo{journal}{\emph{Proceedings of the ACM on Interactive,
  Mobile, Wearable and Ubiquitous Technologies}} \bibinfo{volume}{1},
  \bibinfo{number}{3} (\bibinfo{year}{2017}), \bibinfo{pages}{1--19}.
\newblock


\bibitem[\protect\citeauthoryear{Zhang, Xue, Waghmare, Jain, Pu, Hersek, Lyons,
  Cunefare, Inan, and Abowd}{Zhang et~al\mbox{.}}{2017b}]%
        {zhang2017soundtrak}
\bibfield{author}{\bibinfo{person}{Cheng Zhang}, \bibinfo{person}{Qiuyue Xue},
  \bibinfo{person}{Anandghan Waghmare}, \bibinfo{person}{Sumeet Jain},
  \bibinfo{person}{Yiming Pu}, \bibinfo{person}{Sinan Hersek},
  \bibinfo{person}{Kent Lyons}, \bibinfo{person}{Kenneth~A Cunefare},
  \bibinfo{person}{Omer~T Inan}, {and} \bibinfo{person}{Gregory~D Abowd}.}
  \bibinfo{year}{2017}\natexlab{b}.
\newblock \showarticletitle{Soundtrak: Continuous 3d tracking of a finger using
  active acoustics}.
\newblock \bibinfo{journal}{\emph{Proceedings of the ACM on Interactive,
  Mobile, Wearable and Ubiquitous Technologies}} \bibinfo{volume}{1},
  \bibinfo{number}{2} (\bibinfo{year}{2017}), \bibinfo{pages}{1--25}.
\newblock


\bibitem[\protect\citeauthoryear{Zhang and Sawchuk}{Zhang and Sawchuk}{2012}]%
        {zhang2012usc}
\bibfield{author}{\bibinfo{person}{M. Zhang} {and} \bibinfo{person}{A.
  Sawchuk}.} \bibinfo{year}{2012}\natexlab{}.
\newblock \showarticletitle{USC-HAD: a daily activity dataset for ubiquitous
  activity recognition using wearable sensors}.
\newblock


\bibitem[\protect\citeauthoryear{Zhang, Isola, and Efros}{Zhang
  et~al\mbox{.}}{2016b}]%
        {zhang2016colorful}
\bibfield{author}{\bibinfo{person}{Richard Zhang}, \bibinfo{person}{Phillip
  Isola}, {and} \bibinfo{person}{Alexei~A Efros}.}
  \bibinfo{year}{2016}\natexlab{b}.
\newblock \showarticletitle{Colorful image colorization}. In
  \bibinfo{booktitle}{\emph{European conference on computer vision}}. Springer,
  \bibinfo{pages}{649--666}.
\newblock


\bibitem[\protect\citeauthoryear{Zhang, Zhao, Nguyen, Xu, Sen, Hester, and
  Alshurafa}{Zhang et~al\mbox{.}}{2020}]%
        {zhang2020necksense}
\bibfield{author}{\bibinfo{person}{Shibo Zhang}, \bibinfo{person}{Yuqi Zhao},
  \bibinfo{person}{Dzung~Tri Nguyen}, \bibinfo{person}{Runsheng Xu},
  \bibinfo{person}{Sougata Sen}, \bibinfo{person}{Josiah Hester}, {and}
  \bibinfo{person}{Nabil Alshurafa}.} \bibinfo{year}{2020}\natexlab{}.
\newblock \showarticletitle{NeckSense: A Multi-Sensor Necklace for Detecting
  Eating Activities in Free-Living Conditions}.
\newblock \bibinfo{journal}{\emph{Proceedings of the ACM on Interactive,
  Mobile, Wearable and Ubiquitous Technologies}} \bibinfo{volume}{4},
  \bibinfo{number}{2} (\bibinfo{year}{2020}), \bibinfo{pages}{1--26}.
\newblock


\bibitem[\protect\citeauthoryear{Zhao, Wu, Ye, Guo, and Zhang}{Zhao
  et~al\mbox{.}}{2020}]%
        {zhao2020musicoder}
\bibfield{author}{\bibinfo{person}{Yilun Zhao}, \bibinfo{person}{Xinda Wu},
  \bibinfo{person}{Yuqing Ye}, \bibinfo{person}{Jia Guo}, {and}
  \bibinfo{person}{Kejun Zhang}.} \bibinfo{year}{2020}\natexlab{}.
\newblock \showarticletitle{MusiCoder: A Universal Music-Acoustic Encoder Based
  on Transformers}.
\newblock \bibinfo{journal}{\emph{arXiv preprint arXiv:2008.00781}}
  (\bibinfo{year}{2020}).
\newblock


\end{thebibliography}

\typeout{get arXiv to do 4 passes: Label(s) may have changed. Rerun}

\end{document}